\documentclass[10pt,twocolumn,letterpaper]{article}

\usepackage[pagenumbers]{cvpr} 
\usepackage[shortlabels]{enumitem}
\setlist[itemize]{noitemsep,leftmargin=*,topsep=0em}
\setlist[enumerate]{noitemsep,leftmargin=*,topsep=0em}

\usepackage[dvipsnames]{xcolor}

\definecolor{cvprblue}{rgb}{0.21,0.49,0.74}
\usepackage[pagebackref,breaklinks,colorlinks,citecolor=cvprblue]{hyperref}

\usepackage{calc}
\usepackage{tikz}
\usepackage{booktabs}
\usepackage{balance}
\usepackage{multirow}
\usepackage{overpic}
\usepackage{balance}
\usepackage{cite}
\usepackage[font=small,labelfont=bf]{caption}
\usepackage{color}
\usepackage{pifont}
\makeatletter
\@namedef{ver@everyshi.sty}{}
\makeatother
\usepackage{tikz}
\usetikzlibrary{patterns}

\usepackage{colortbl}
\usepackage{pgfplots}
\pgfplotsset{compat=1.17}
\usepackage{tabularx}
\usepackage{arydshln}
\usepackage{amssymb}
\usepackage{pifont}

 \usepackage{amsmath}
\usepackage{MnSymbol}%
\usepackage{wasysym}%
\usepackage{enumitem}
\usepackage{wrapfig}
\usepackage{xspace}
\usepackage{tabu}
\usepackage[super]{nth}
\usepackage{scalefnt}
\usepackage{dsfont}
\usepackage{graphicx,calc}
\usepackage{algorithm}
\usepackage{algpseudocode}

\definecolor{darkgreen}{RGB}{0,153,51}
\definecolor{linkgreen}{RGB}{52,130,48}
\definecolor{LightCyan}{rgb}{0.87,0.92,0.96}
\definecolor{m_green}{RGB}{233, 254, 187}
\definecolor{m_orange}{RGB}{255, 212, 121}
\definecolor{m_red}{RGB}{255, 190, 188}
\definecolor{m_violet}{RGB}{215, 131, 255}
\definecolor{m_blue}{RGB}{186, 234, 255}
\definecolor{m_brown}{RGB}{255,212,120}
\definecolor{m_lightgreen}{RGB}{212,251,122}
\definecolor{notetext}{rgb}{0.7,0,0}

\definecolor{model_pink}{RGB}{235, 106, 164}
\definecolor{model_orange}{RGB}{250, 194, 122}
\definecolor{model_green}{RGB}{164, 210, 162}
\definecolor{model_gray}{RGB}{120, 120, 120}
\definecolor{model_yellow}{RGB}{251, 231, 171}
\definecolor{model_purple}{RGB}{190, 146, 211}

\usepackage{amssymb}
\usepackage{pifont}
\def\eg{\emph{e.g.}\@\xspace}

\newcommand{\parag}[1]{\vspace{-4px} \vskip8pt \noindent \textbf{#1}}

\newcommand{\name}{{LabelMaker}}

\newcolumntype{Y}{>{\centering\arraybackslash}X}
\newcolumntype{Z}{>{\raggedleft\arraybackslash}X}

\usepackage{array}
\newcolumntype{P}[1]{>{\centering\arraybackslash}p{#1}}
\newcolumntype{M}[1]{>{\centering\arraybackslash}m{#1}}

\definecolor{darkblue}{RGB}{60, 82, 145}
\definecolor{kingblue}{RGB}{65, 105, 225}

\definecolor{background}{RGB}{226, 226, 226}
\definecolor{head}{RGB}{210, 78, 142}
\definecolor{rightArm}{RGB}{255, 176, 0}
\definecolor{leftArm}{RGB}{228, 162, 227}
\definecolor{rightForeArm}{RGB}{90, 64, 210}
\definecolor{leftForeArm}{RGB}{243, 232, 88}
\definecolor{rightHand}{RGB}{158, 143, 20}
\definecolor{leftHand}{RGB}{192, 100, 119}
\definecolor{torso}{RGB}{100, 143, 255}
\definecolor{hips}{RGB}{129, 103, 106}
\definecolor{rightUpLeg}{RGB}{243, 115, 68}
\definecolor{leftUpLeg}{RGB}{152, 200, 156}
\definecolor{rightLeg}{RGB}{149, 192, 228}
\definecolor{leftLeg}{RGB}{152, 78, 163}
\definecolor{rightFoot}{RGB}{129, 0, 50}
\definecolor{leftFoot}{RGB}{76, 134, 26}

\newlength\myheight
\newlength\mydepth
\settototalheight\myheight{Xygp}
\usetikzlibrary{positioning}

\def\paperID{4} 
\def\confName{3DV\xspace}
\def\confYear{2024\xspace}

\title{
\vspace{-32px}
\textsc{\name}:\\Automatic Semantic Label Generation from RGB-D Trajectories
\vspace{-15px}
}

\author{
Silvan Weder$^{*, 1}$ \quad
Hermann Blum$^{*, 1}$ \quad
Francis Engelmann$^{1, 2, 3}$ \quad
Marc Pollefeys$^{1, 4}$
\\ \vspace{-7px}\\
$^{1}$ETH Zurich \quad $^{2}$ETH AI Center \quad $^{3}$Google \quad $^{4}$Microsoft
\\ \vspace{-7px}\\
}

\begin{document}
\twocolumn[{%
\renewcommand\twocolumn[1][]{#1}%
\maketitle
\thispagestyle{empty}
\vspace{-15px}
\includegraphics[width=1.0\linewidth, trim={0 0 0 0cm},clip]{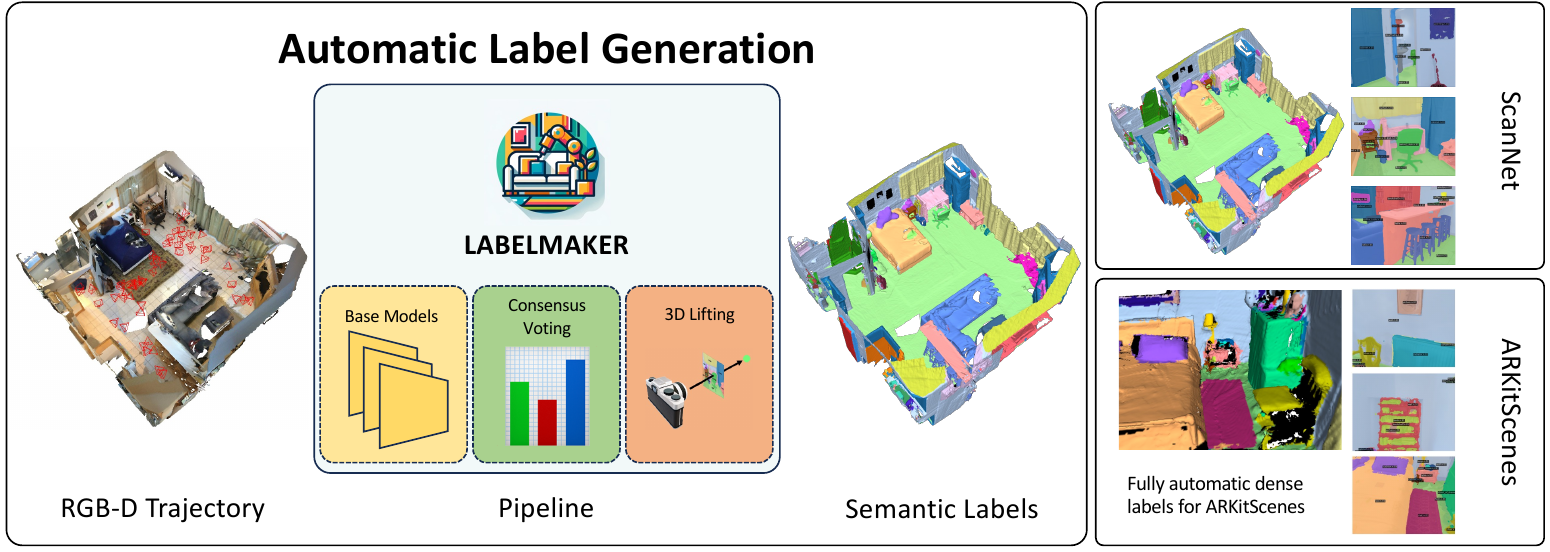}
\vspace{-18px}
\captionof{figure}{
\name{} bundles a collection of state-of-the-art segmentation models with different sets of predicted classes in a neural field.
\name{} can refine existing annotations and produce highly accurate 2D as well as 3D labels on ScanNet \emph{(top-right)}.
At the same time, it opens new possibilities to rapidly label large-scale datasets without human effort such as ARKitScenes \emph{(bottom-right)}.
}

\label{fig:teaser}
\vspace{0.5cm}
}]
\begin{abstract}
Semantic annotations are indispensable to train or evaluate perception models, yet very costly to acquire.
This work introduces a fully automated 2D/3D labeling framework that, without any human intervention, can generate labels for RGB-D scans at equal (or better) level of accuracy than comparable manually annotated datasets such as ScanNet.
Our approach is based on an ensemble of state-of-the-art segmentation models and 3D lifting through neural rendering.
We demonstrate the effectiveness of our \name{} pipeline by generating significantly better labels for the ScanNet datasets and automatically labelling the previously unlabeled ARKitScenes dataset. Code and models are available at \href{https://labelmaker.org/}{labelmaker.org}.
\end{abstract}

\section{Introduction}
Semantic perception of the world around us is of central importance for many computer vision applications \cite{mask3d, takmaz20233d, yue2023connecting}. 
Without semantic perception, meaningful interactions with our environment are hardly possible. 
Thus, semantic scene perception has been a long-standing problem in computer vision and robotics \cite{nekrasov2021mix3d, mask3d, kreuzberg20224d, chibane2022box2mask}.
In recent years, most solutions have converged towards using deep neural networks.
However, training and evaluating these networks is hard.
As recent works such as SAM~\cite{sam}, language-based models~\cite{ovseg,openscene,takmaz2023openmask3d}, or InternImage~\cite{wang2022internimage} have shown, huge quantities of training data, orders of magnitude larger than any single existing research dataset, are necessary to achieve good generalization. 
On the other hand, generalization is necessary because the distribution of the deployment environment - \eg{}, a particular user's home, in which a robotic application is to be deployed  - is outside of the distribution of existing annotated training datasets.
To evaluate generalization in or adapt to specific deployment environments, labeled data of these environments is required.
From both training and deployment perspectives, the availability of labeled data is therefore a key problem. 
Unfortunately, the acquisition of this data is usually very expensive as semantic ground-truth annotation is a time-consuming manual process.

In this work, we particularly focus on 3D semantic segmentation.
The available scale of 3D semantic segmentation data such as ScanNet~\cite{scannet} or Matterport3D~\cite{matterport} is far below the scale of 2D semantic segmentation datasets like ADE20k~\cite{ade20k}, COCO-stuff~\cite{caesar2018coco}, or others~\cite{silberman_indoor_2012,yu2020bdd100k,cordts_cityscapes_2016}.
Even tough tasks such as semantic segmentation or online semantic reconstruction gain maturity and are crucial for interactive applications, there is even less semantic data with paired camera trajectories and corresponding scene reconstructions.
ScanNet~\cite{scannet} is by far the largest in this domain with an abundance of scenes and a well-established benchmark.
However, both camera images and labels are oftentimes noisy, making it hard to generalize from ScanNet to other datasets.
ARKitScenes~\cite{dehghan2021arkitscenes} shows the growing possibility to capture RGB-D trajectories at scale, and at the same time illustrates the cost of semantic annotations, featuring an incomplete list of bounding boxes. 

To push the scale and accuracy of 3D semantic segmentation datasets, we present \emph{\name}. \name\ automatically creates labels that are on the same level of accuracy as the established ScanNet benchmark, but without any human annotation.
Further, we show that it can produce better labels than the original ScanNet labels when using the human annotations as an additional input.

The design of our method is motivated by two observations. 
The first observation is on recent advances in 2D semantic segmentation, where a leap in training data scale through combination of different tasks and datasets~\cite{wang2022internimage} or visual-language models~\cite{ovseg} has boosted generalization.
The second observation is in the field of neural radiance fields, where~\cite{zhi2021place,Liu2023Unsupervised,siddiqui2023panoptic} have shown that NeRFs can be used to denoise semantic input labels and learn a multi-view consistent semantic label field.
We leverage these two observations and motivate an automatic labelling pipeline with two main components at its heart.
First, we leverage large 2D models, that combine the power of different tasks and input modalities, in order to predict different hypothesis for labels in 2D. 
These labels are aggregated using our consensus voter in order to obtain a single 2D prediction for every frame.
Second, all 2D predictions are aggregated and made consistent using a neural radiance field.
This neural radiance field can be used to render clean and consistent 2D label maps. Alternatively, the labels can be aggregated and mapped into 3D to obtain labeled pointclouds or meshes.

With a comparison to SOTA methods and datasets and an extensive ablation study, we showcase that our method automatically generates labels of similar quality than human annotators. We also demonstrate fully automatic labelling for ARKitScenes, for which no dense labels exist to date.

In summary, our contributions are:
\begin{itemize}
    \item A curated mapping between the indoor label sets NYU40, ADE20k, ScanNet, Replica, and into the wordnet graph.
    \item A pipeline to automatically label RGB-D trajectories, as well as corresponding 3D point clouds, that achieves higher quality than the original labels of ScanNet.
    \item Generated labels in 3D meshes and 2D images for ScanNet \cite{scannet} and ARKitScenes \cite{dehghan2021arkitscenes}.
\end{itemize}

\section{Related Work}

\parag{Labelling in 2D.} Cityscapes~\cite{cordts_cityscapes_2016} is one of the most established 2D semantic segmentation datasets.
The authors report an effort of more than 1.5h to annotate a single frame. 
Similar frame-by-frame manual annotations were provided in NYU Depth~\cite{silberman_indoor_2012}, ADE20k~\cite{ade20k}, or COCO-stuff~\cite{caesar2018coco}.
While frame-by-frame annotations yield very high quality segmentation masks, they are expensive to obtain. 
Although the effort can be reduced through comfortable annotation tools~\cite{labelme,Breheret:2017}, it cannot be avoided that a human inspects every image and performs at least a couple of clicks.

\parag{Labelling in 3D.} If scenes are annotated in 3D, their annotations can easily be rendered into any localized camera image in the same scene, therefore potentially reducing labeling effort.
This approach was followed in Replica~\cite{replica} and ScanNet~\cite{scannet}. iLabel~\cite{zhi2021ilabel} pioneered to use NeRFs for this type of rendering, additionally showing that NeRFs have an intrinsic capability to segment whole objects along texture boundaries from a few clicks. Similarly,~\cite{kontogianni2023interactive, yue2023agile3d} also reduce the manual labelling effort to a few positive and negative clicks per object.
Matterport~\cite{matterport} consists of large labeled 3D scans, but does not have corresponding 2D images and therefore can only be used for 3D methods.

\parag{Pretrained Models.} It is a well-established approach in labelling to label parts of a dataset, train a model on that part, and use its predictions to bootstrap labels for the rest of the data. More recently, models pretrained on large amounts of data have been introduced to help labelling completely unseen datasets. SAM~\cite{sam} showed impressive results of segmenting objects in images from close to zero clicks where only labels have to be assigned. The seconds step can even be bootstrapped through CLIP~\cite{clip}. CLIP2Scene~\cite{chen2023clip2scene} takes a similar approach in 3D to train a pointcloud classifier on previously unlabeled data.

\section{Method}
We briefly discuss the relabelling of ScanNet scenes. 
Then, we discuss the translation between prediction spaces.
Finally, we present our automatic labelling pipeline.

\subsection{Relabeling ScanNet Scenes}
\label{subs:relabeling}
To be able to evaluate the quality of \name{}, we want to compare it against existing human annotations. 
We choose the ScanNet dataset because its scale has a large potential for automatic processing. 
To be able to evaluate the quality of the existing labels and compare them with \name, we create high-quality annotations for a selection of scenes.

The original ScanNet~\cite{scannet} labels were created using free text user prompts. They consequently have duplicates or are ill-defined. This reflects the open-world approach of \citet{scannet}, but contradicts the use as benchmark labels, for which they map them to other class sets. As a set of annotation classes, we therefore did not directly annotate with ScanNet classes, but use wordnet~\cite{miller1995wordnet} synkeys\footnote{Wordnet is a dictionary and synkeys are the names of its entries. I.e., a set of synonymous words has 1 synkey, but a word with different meanings as one synkey per definition.}.
In particular, we start from the mapping that ScanNet defined between their labels and wordnet and take the categories that occur at least three times in the dataset. 
This yields an initial list of 199 categories, already resolving many ambiguities. 
We then check the definitions of all of these categories in the wordnet database and correct the initial mapping, as well as merged categories that are still too ambiguous by their definitions in wordnet (e.g. \texttt{rug.n.01} ``rug, carpet, carpeting; floor covering consisting of a piece of thick heavy fabric (usually with nap or pile)'' and \texttt{mat.n.01} ``a thick flat pad used as a floor covering'' ). The result are 186 categories that come with a text definition, a defined hierarchy, and all possible synonyms that describe the category.

We then annotate our selected ScanNet scenes with these 186 categories based on their wordnet definitions. We use~\cite{kontogianni2023interactive} to annotate the fine meshes of the scenes with a minimum number of necessary clicks. Only the authors of this paper provided annotations, and each annotation was cross-checked by at least one other author. In case of doubt, individual objects were discussed together. On average, labeling of a scene took 5 hours.

\subsection{Translation between Prediction Spaces}
\label{subs:label-translation}
We employ different predictors that were trained on different data sets with different numbers and definitions of classes. This requires translating between different prediction spaces. We therefore build a mapping between the class definitions of NYU40, ADE20k, ScanNet20, ScanNet200, Replica, and the WordNet semantic language graph.

In this effort, we build on top of previous work, as the original ScanNet~\cite{scannet} already defined a mapping between ScanNet classes, NYU40 classes, Eigen13 classes, and wordnet synkeys~\cite{miller1995wordnet}. Furthermore, \citet{lambert_mseg_2020} curated mappings between the taxonomies of semantic segmentation datasets, out of which NYU40, SUNRGBD, and ADE20k are most relevant for indoor perception. 
We take the union of both works as initial mapping, but find that many corrections are needed, especially with regard to wordnet synkeys, and many ADE20k are missing because \cite{lambert_mseg_2020} only considered 20 out of 40 NYU categories. We then add mappings to the Replica categories for the purpose of evaluation, since Replica is one of the most accurately annotated indoor semantic datasets.

When mapping between two class spaces,
for any class in the source space there are three cases in the target space:
\emph{a)} there is no corresponding class in the target space,
\emph{b)} there is exactly one corresponding class in the target space.
This may be an exact match,
or a class to which multiple class ids from the source space are matched
(\eg{}, the source space may distinguish between
office chair, chair, and stool
but the target space just has one general chair class),
\emph{c)} there are multiple corresponding classes in the target space because the target space has a higher resolution than the source space
(\eg{}, a general chair class in the source space can be split up in the target space to distinguish between office chair, chair, or stool).

   
For (a) and (b), mappings are straightforward. We resolve (c) dependent on the use cases: \begin{itemize}
    \item \emph{Evaluating a class with multiple correspondences.} A label of any of the correspondences is treated as a true positive. If none of the correspondences is the true class, all of them are counted as false positives.
    \item \emph{Computing model consensus.} 
    Predictions in the source space vote for all possible correspondences in the target space. 
    The ambiguity between the possible correspondences is usually resolved through an additional predictor with a prediction space of higher resolution. 
    If no resolution is achieved, we pick the first of the possible classes.
\end{itemize}

\begin{figure*}[ht!]
    \centering
    \includegraphics[width=\textwidth]{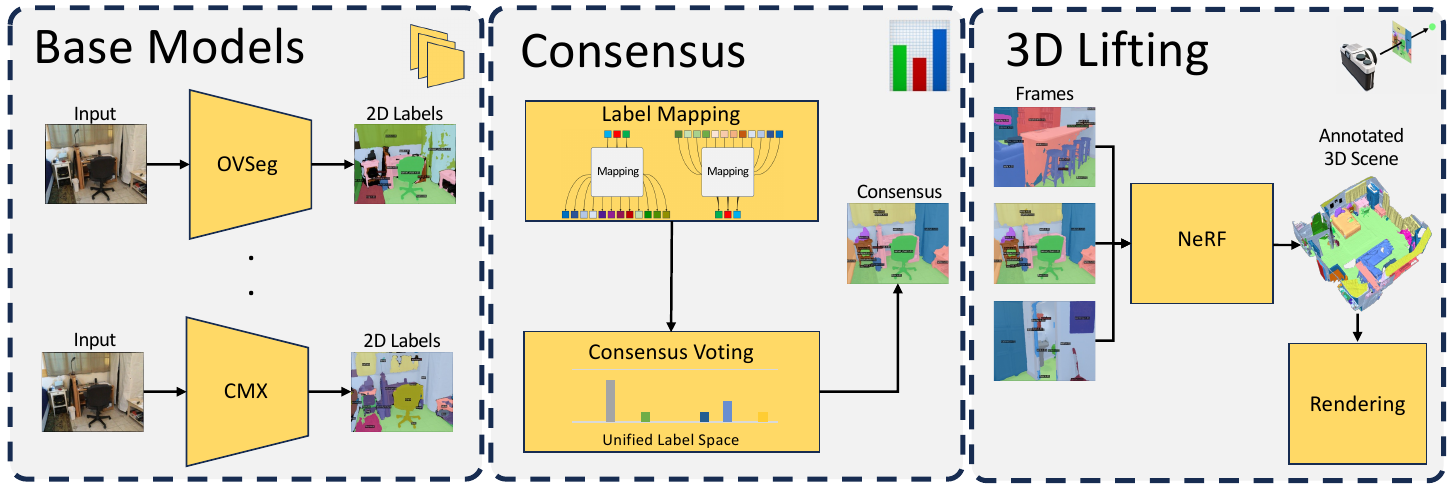}
\caption{\textbf{Pipeline Overview.} The base models predict individual semantic maps for each 2D frame of the trajecotry. The consensus first maps the label spaces in our unified label space and then runs our consensus voting mechanism for every frame. Finally, the 3D lifting aggregates the per-frame predictions in 3D that improves the segment quality due to the additional denoising. The final 3D annotation can be rendered back into 2D to obtain a multi-view consistent labelling across the entire trajecotry.
\vspace{-12px}}
\label{fig:pipeline}
\end{figure*}

\subsection{Base Models}
\label{subs:models}
We employ an ensemble of strong base models, each state-of-the-art in their respective task and data characteristic:


InternImage~\cite{wang2022internimage} is a supervised 2D RGB-only semantic segmentation model that at the time of writing has state-of-the-art performance on the Cityscapes and ADE20k benchmarks. It achieves this by performing large-scale joined pretraining on most available visual classification datasets. We use the ADE20k fine-tuned variant.

OVSeg~\cite{liang2023open} is an open-vocabulary semantic segmentation model based on CLIP~\cite{clip}, a visual-language representation model. OVSeg segments images by assigning region proposals to a set of given prompts and is therefore not limited to a fixed set of classes. In particular, we added such an open-vocabulary segmentation model not because they achieve the best performance on a given task but because of their generalization ability. We generate prompts from our set of wordnet synkeys by averaging over language prompts such as ``A \_ in a room.'', but also using all possible synonyms according to wordnet.

CMX~\cite{zhang2023cmx} is at the time of writing the state-of-the-art 2D semantic segmentation model for NYU Depth v2, a RGB+Depth indoor dataset. Its predictions also take the geometric cues from the depth into account.

Mask3D~\cite{mask3d} is at the time of writing the state-of-the-art 3D instance segmentation model on ScanNet200~\cite{scannet200}. This method operates on an accumulated pointcloud of a scene instead of frames, therefore taking the geometry even better into account. It is trained on ScanNet.
We render the 3D semantic instance predictions into the 2D training frames to map them into the same space as all other base models.

The four semantic models produce classifications in four different sets of classes. InternImage predicts 150 ADE20k classes, CMX predicts 40 NYU classes, Mask3D predicts 200 ScanNet classes, and our OVSeg prompts cover 186 wordnet classes. In addition to the semantic models, we use OmniData~\cite{eftekhar2021omnidata} to complement the depth sensor.

\subsection{Model Consensus}
As illustrated in Fig.~\ref{fig:pipeline},
we run all models of Sec.~\ref{subs:models} individually on every frame and then,
per frame, merge their predictions together using the translation described in Sec.~\ref{subs:label-translation}.
We further use left-right flipping as test time augmentation,
which means that each pixel receives votes for possible classes from:
\begin{itemize}
    \item the standard RGB image and it's flipped version for the 2D segmentation models InternImage, CMX, and OVSeg
    \item 2 votes (to equalize the test-time augmentation of the RGB frame) from the Mask3D prediction rendered into the current frame
    \item in the variant where we additionally use available human annotations, 5 votes from the original ScanNet labels
\end{itemize}
For every pixel, we choose the class with the maximum number of votes.
If no class has sufficient votes (parameterized as a threshold),
we set the prediction to ``\emph{unknown}" and it will have no loss in the 3D lifting.

\subsection{3D Lifting}
By computing a consensus over a diverse set of 2D predictors, we leverage the knowledge and scale of 2D semantic segmentation datasets.
However, the per-frame predictions are noisy and often inconsistent, especially around image boundaries. 
These inconsistencies can be mitigated and the performance can even be improved, as previous work has shown~\cite{siddiqui2023panoptic,Liu2023Unsupervised}, by lifting the 2D predictions into 3D.

Therefore, we leverage the recent progress based on NeRFs to generate multi-view consistent 2D semantic segmentation labels in all frames. 
Based on the observation in previous works~\cite{siddiqui2023panoptic,Liu2023Unsupervised} that accurate geometry is important to resolve inconsistencies between predictions of multiple frames instead of hallucinating geometry that would explain semantic predictions, we train an implicit surface model from sdfstudio~\cite{sdfstudio} that has a more explicit surface definition compared to a NeRF yielding improved geometry compared to vanilla NeRF. 
Thus, we add a semantic head to the Neus-Acc model, train it on all views with losses from RGB reconstruction, sensor depth, monocular normal estimation, and our semantic consensus. 
Finally, we render the optimized semantics back into all camera frames.

To generate consistent 3D semantic segmentation labels, we follow an established and more direct approach. 
Given a pointcloud of the scene, we project the pointcloud into each consensus frame to find corresponding pixels and then take a majority vote over all pixels corresponding to a point.

\section{Experiments}
\begin{table*}[]
\setlength{\tabcolsep}{3pt}
    \centering
    \small
    \begin{tabular}{r cccccc cccccc}
    \toprule
    & \multicolumn{6}{c}{2D} & \multicolumn{6}{c}{3D}\\
    \cmidrule(r){2-7}  \cmidrule(r){8-13}
    evaluation class set & \multicolumn{3}{c}{\textbf{NYU} \textit{(40 classes)}} & \multicolumn{3}{c}{\textbf{wordnet} \textit{(186 classes)}} & \multicolumn{3}{c}{\textbf{NYU} \textit{(40 classes)}} & \multicolumn{3}{c}{\textbf{wordnet} \textit{(186 classes)}}\\
    metric
    &  mIoU & mAcc & tAcc
    &  mIoU & mAcc & tAcc
    &  mIoU & mAcc & tAcc
    &  mIoU & mAcc & tAcc\\
                    \midrule
        ScanNet labels~\cite{scannet} & 47.7 & 56.2 & 69.2 & 38.1 & 46.3 & 69.7 & 40.1 & 48.2 & 68.6 & 17.7 & 21.3  & 70.6 \\
        SemanticNerf* \cite{zhi2021place} & 45.2 & 56.6 & 69.3 & 32.9 & 43.7 & 71.2 & 36.7 & 47.1 & 68.4 & 14.8 & 19.3 & 71.0\\
        \name ~w/o ScanNet (automatic labels)  & 50.7 & 64.0 & 75.3 & 33.5 & 43.5 & 72.3 & 41.3 & 47.3 & 71.2 & 15.7 & 18.1 & 71.5\\
        \name{} (Ours) & \textbf{53.4} & \textbf{65.0} & \textbf{77.5} & \textbf{39.1} & \textbf{49.3} & \textbf{77.2} & \textbf{44.1} & \textbf{53.4} & \textbf{76.1} & \textbf{18.2} & \textbf{22.0} & \textbf{76.7}\\
        \bottomrule
    \end{tabular}  
    \vspace{-5px}
    \caption{Comparison of the label quality of the ScanNet labels, \name\ without any human input, and \name\ taking the ScanNet annotations as additional input. The results are measured over 5 scenes from ScanNet against newly annotated high-quality ground truth. Based on our translation of prediction spaces, we measure metrics over the medium-tail NYU40 set of categories and our full long-tail ground truth categories. 
    For NYU40 classes, \name\ is capable of producing labels of higher quality than the ScanNet human annotations, without any human input. For more long-tail categories, the automatic mode does not reach the quality of ScanNet, but \name\ is able to considerably improve human annotations.
    }
    \label{tab:main_table}
\end{table*}

\subsection{Implementation Details}
For the 2D models, we use the corresponding available open-source code and adjust it to our pipeline.
As described in Sec.~\ref{subs:label-translation}, we generate votes from each 2D model into a common label space. We choose our defined 186 class wordnet label space as output. We choose the label with highest votes, but require a minimum of 3 out of 13 (with ScanNet annotations) resp. 4 out of 8 (automatic pipeline) votes.
For 3D optimization, we build on top of SDFStudio~\cite{sdfstudio}, specifically the Neus-Acc~\cite{wang2021neus} model, and add a semantic head and semantic rendering similar to \cite{zhi2021place}. 

\subsection{Datasets}
We run our proposed method on three different datasets to show its performance and validate our design choices.

\parag{ScanNet~\cite{scannet}}
We randomly select 5 scenes from the ScanNet that cover all frequent room types. 
We carefully annotate high-resolution meshes of the scenes using~\cite{kontogianni2023interactive} as described in Sec.~\ref{subs:relabeling} in order to have a complete and accurate groundtruth to evaluate against. 

\parag{Replica~\cite{replica}}
We also evaluate our method on the Replica dataset. 
This is a semi-synthetic dataset, captured as a high accuracy mesh from real environments and then rendered into trajectories in~\cite{zhi2021place}.
We select the 3 `room' scenes and evaluate against the given annotation.

\parag{ARKitScenes~\cite{dehghan2021arkitscenes}}
To showcase the automatic labelling pipeline on an existing dataset, we run it on selected scenes of the ARKitScenes dataset, where only sparse bounding box labels are available up to date. ARKit Scenes consists of trajectories captured with consumer smartphones which are registered to a professional 3D scanner.

\subsection{Baselines}
We mainly compare \name\ to the existing manually created annotations in ScanNet~\cite{scannet}.
As an additional baseline, we report the result of fitting and rendering the ScanNet annotations with our adapted SemanticNeRF~\cite{zhi2021place}. 

\parag{ScanNet~\cite{scannet}.}
For this baseline, we measure the quality of the annotations in ScanNet. 
To this end, we take the raw ScanNet labels and map them into our labelspace defined by wordnet. The mapping from ScanNet IDs to wordnet synkeys is to a large extent already provided in ~\cite{scannet}.

\parag{SemanticNeRF~\cite{zhi2021place}.} This baseline is inspired by~\cite{zhi2021place} and adapted to our pipeline by integrating the semantic head into SDFStudio.
Then, we run this version of SemanticNeRF on the ScanNet 2D semantic labels. 
Thus, we can measure the effect of multi-view aggregation and optimization on the groundtruth ScanNet labels. The hypothesised effect is that through the extra RGB and geometry information provided to the NeRF, segmentation boundaries may be smoother than those of the ScanNet `supervoxels'.

\begin{figure*}
\setlength{\tabcolsep}{1px}%
\newcommand{\lnw}{0.19}
\footnotesize
\centering
\begin{tabular}{ccccc}
    RGB & ScanNet label & \name{}  & \name{} - NYU40 & Ground-truth \\
    \includegraphics[width=\lnw\linewidth]{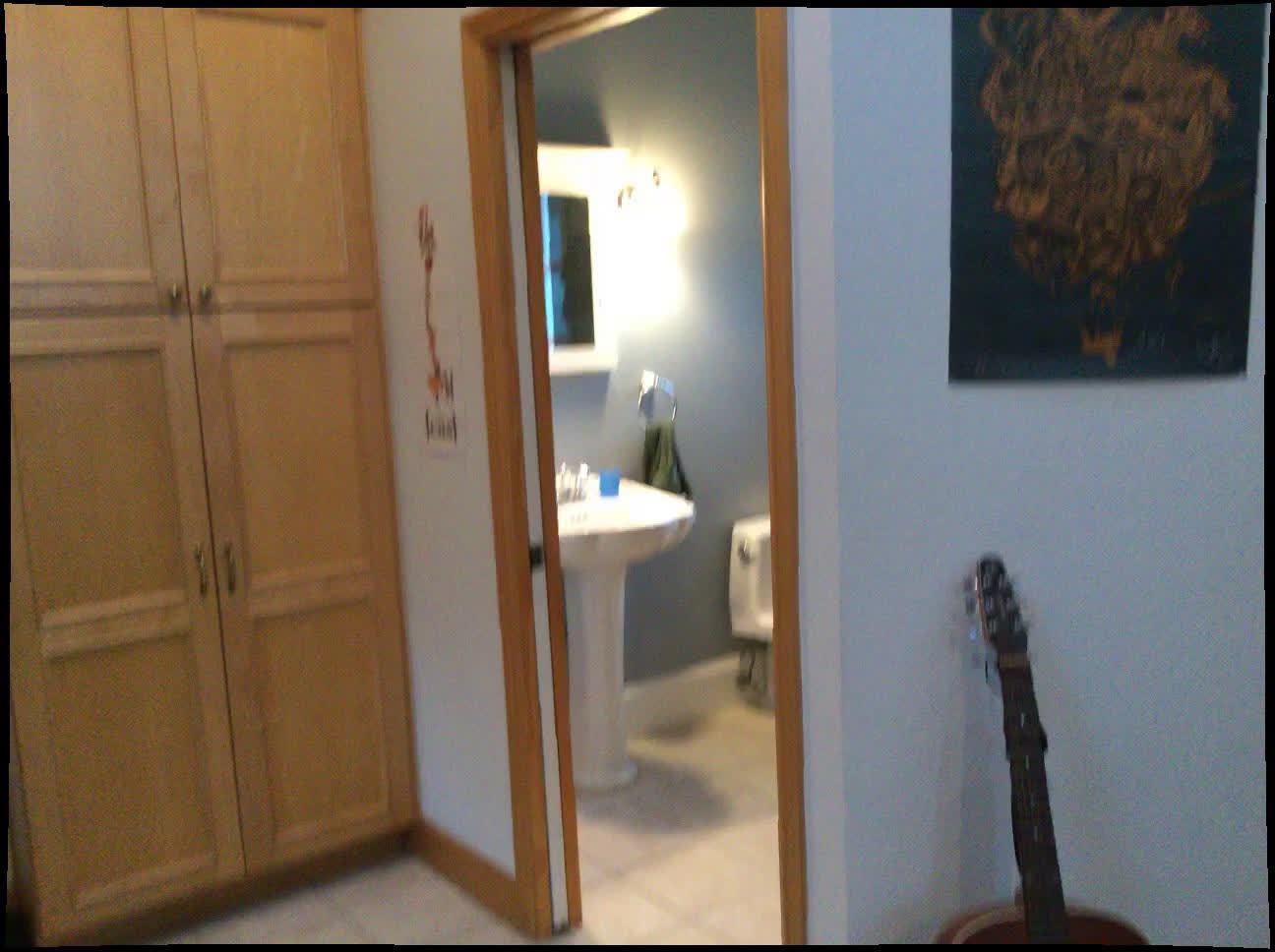} &
    \includegraphics[width=\lnw\linewidth]{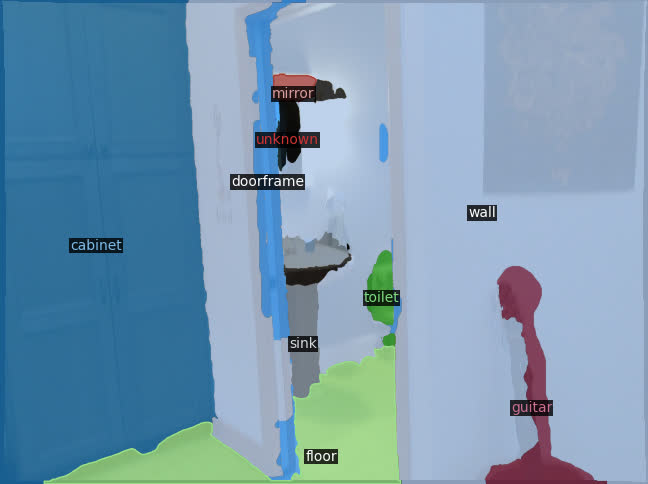} &
    \includegraphics[width=\lnw\linewidth]{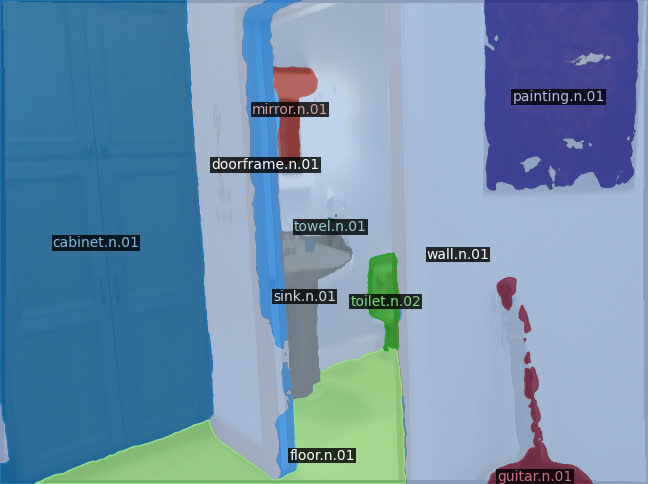} &
    \includegraphics[width=\lnw\linewidth]{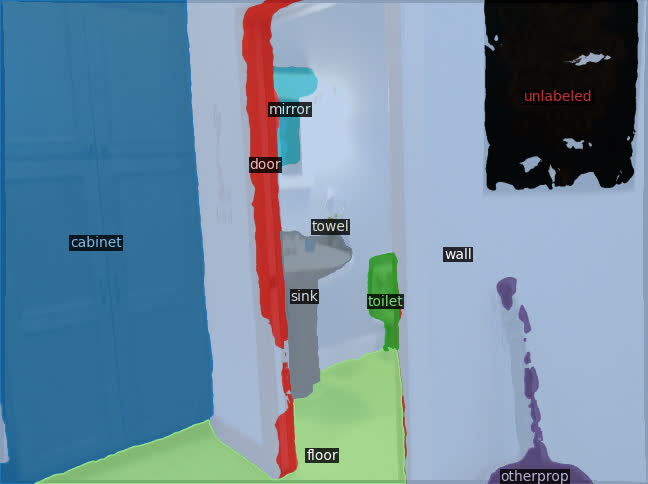} &
    \includegraphics[width=\lnw\linewidth]{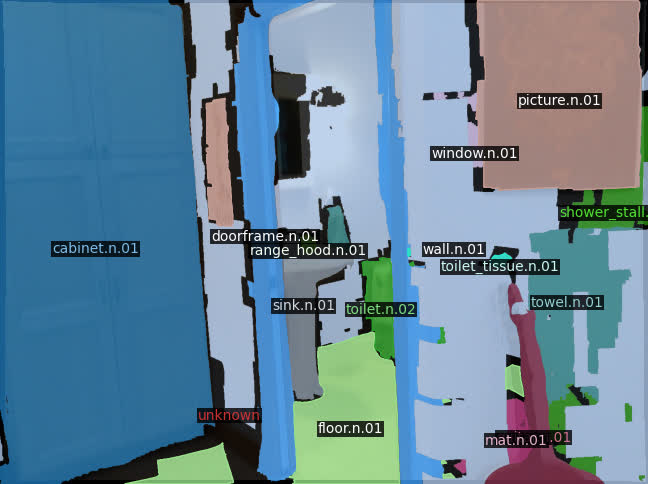} \\
    \includegraphics[width=\lnw\linewidth]{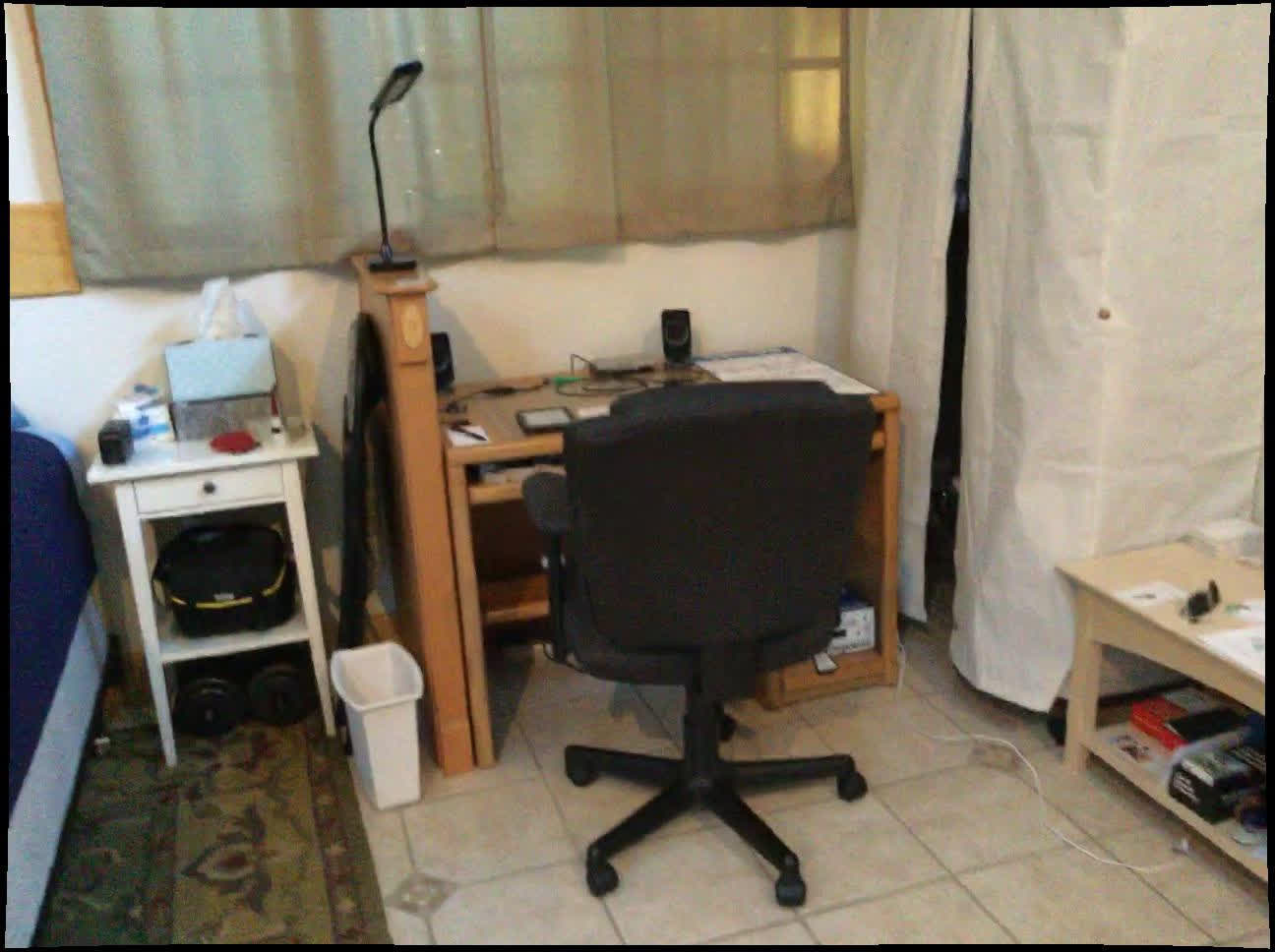} &
    \includegraphics[width=\lnw\linewidth]{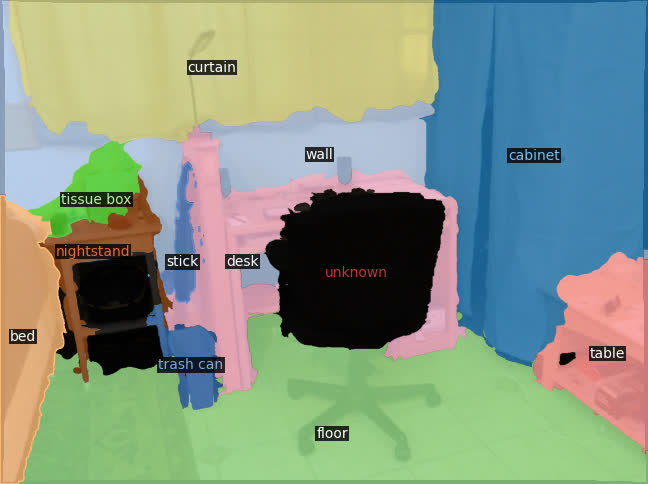} &
    \includegraphics[width=\lnw\linewidth]{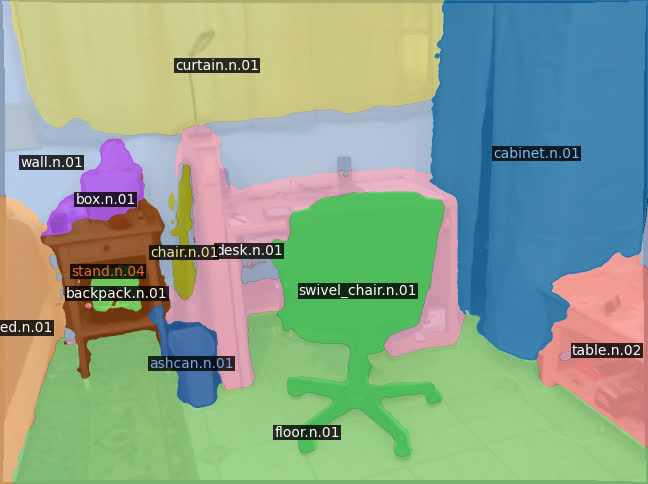} &
    \includegraphics[width=\lnw\linewidth]{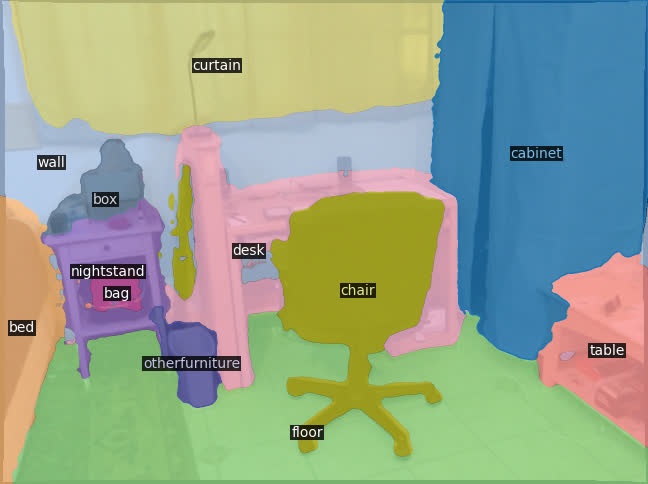} &
    \includegraphics[width=\lnw\linewidth]{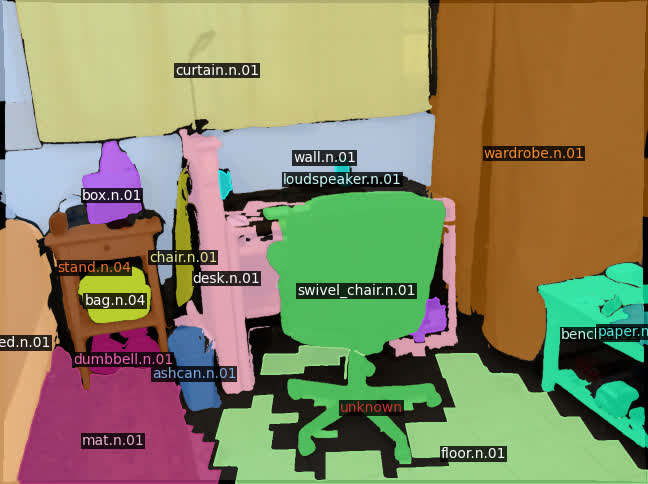} \\
    \includegraphics[width=\lnw\linewidth]{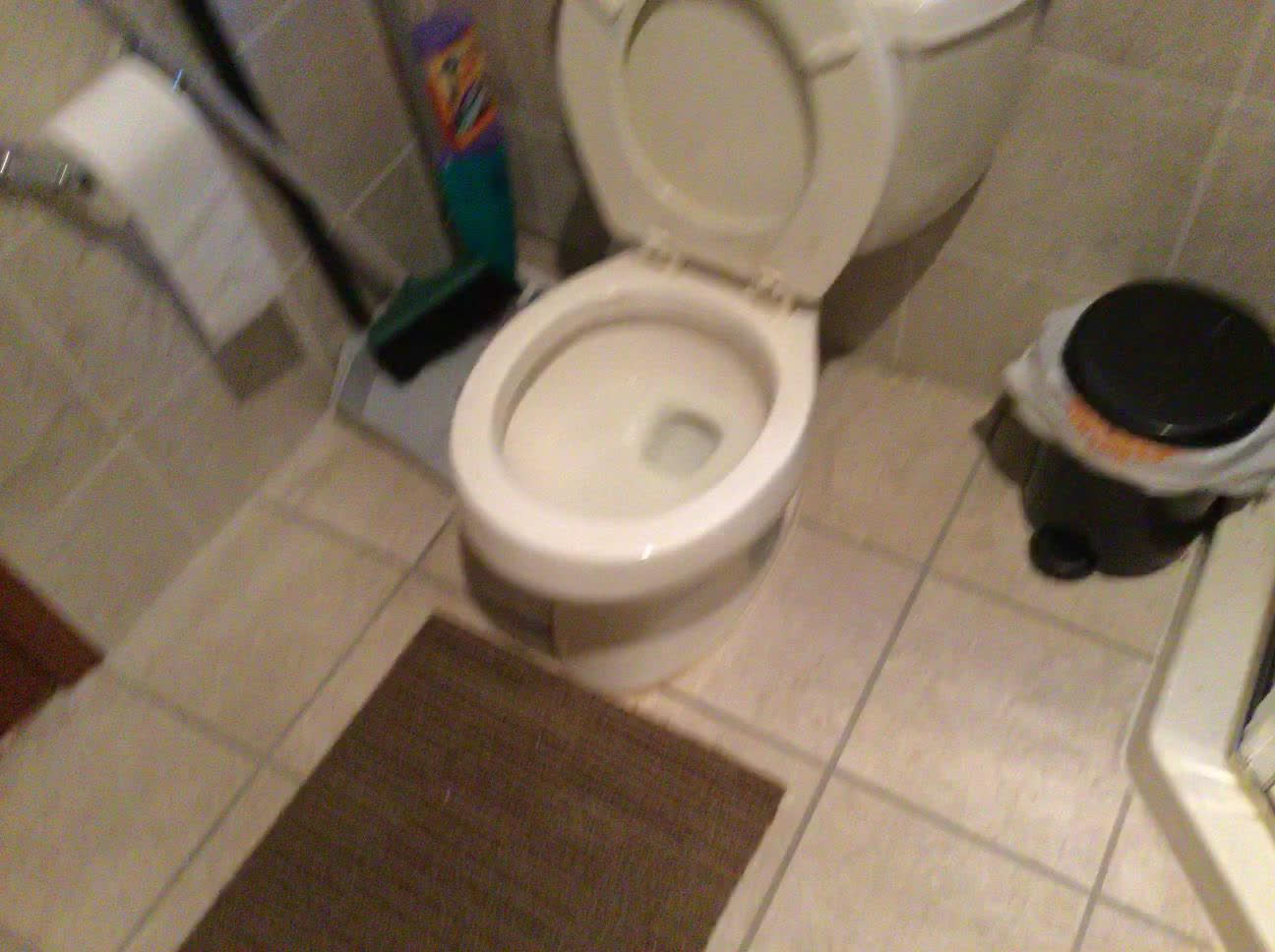} &
    \includegraphics[width=\lnw\linewidth]{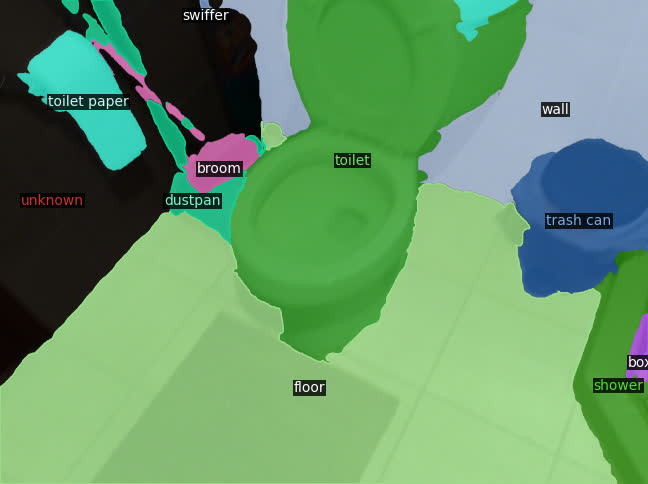} &
    \includegraphics[width=\lnw\linewidth]{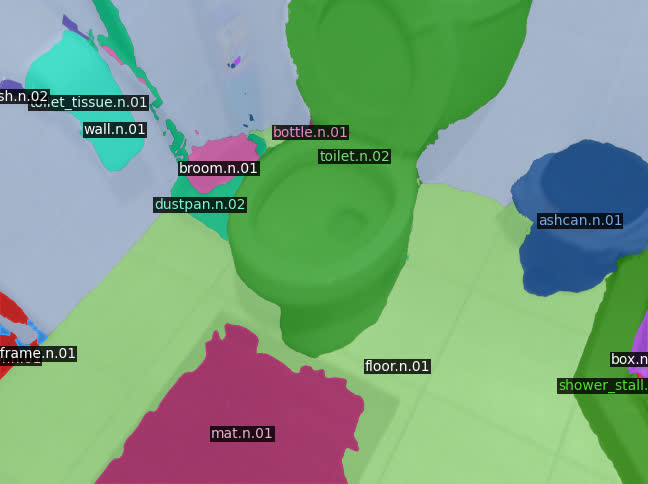} &
    \includegraphics[width=\lnw\linewidth]{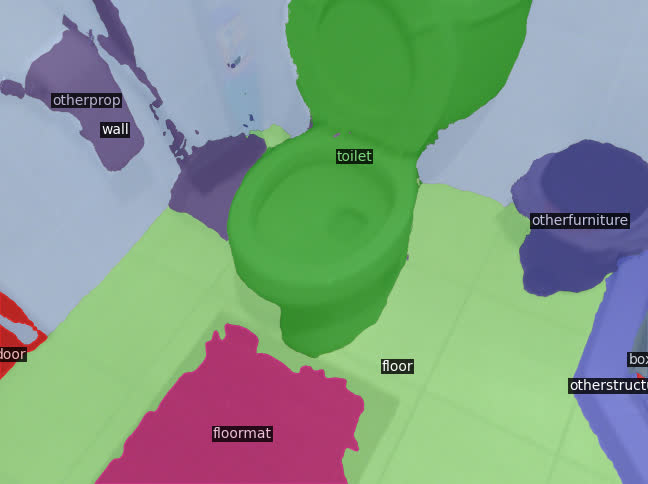} &
    \includegraphics[width=\lnw\linewidth]{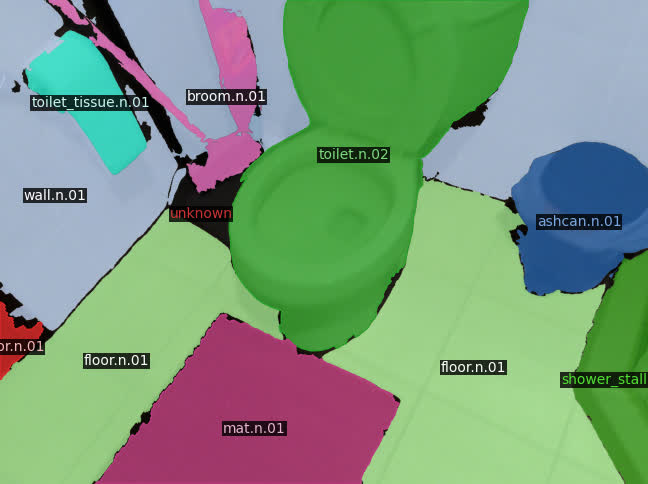} \\
    \includegraphics[width=\lnw\linewidth]{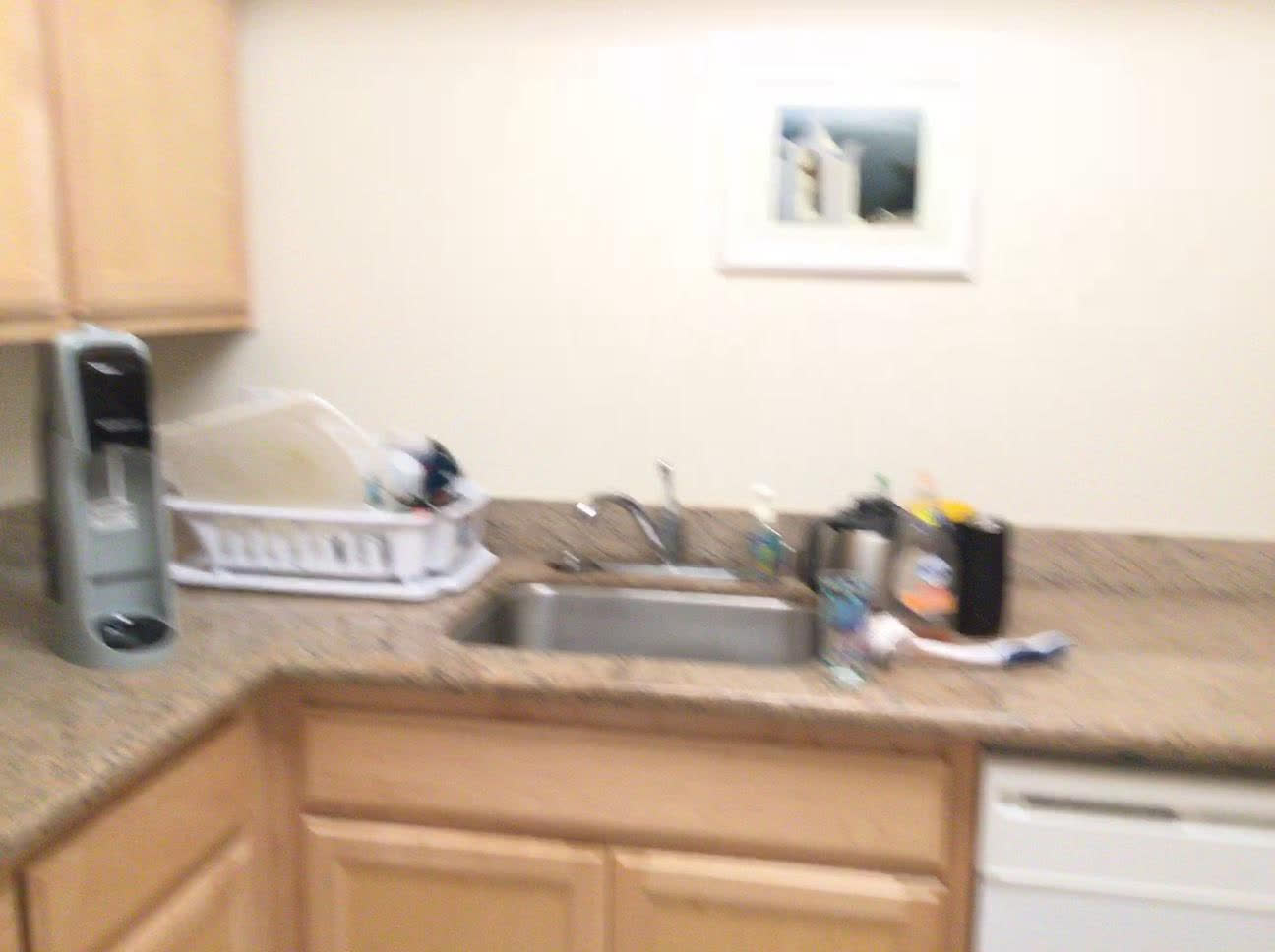} &
    \includegraphics[width=\lnw\linewidth]{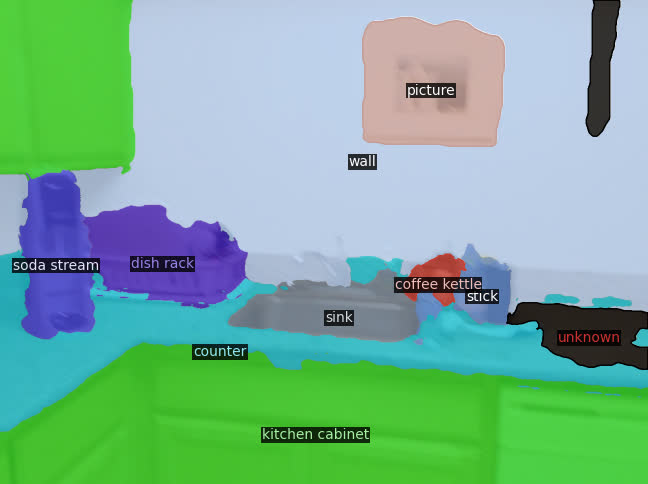} &
    \includegraphics[width=\lnw\linewidth]{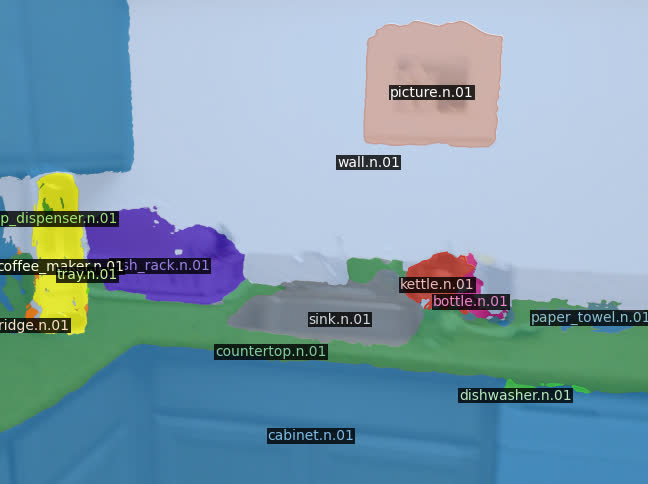} &
    \includegraphics[width=\lnw\linewidth]{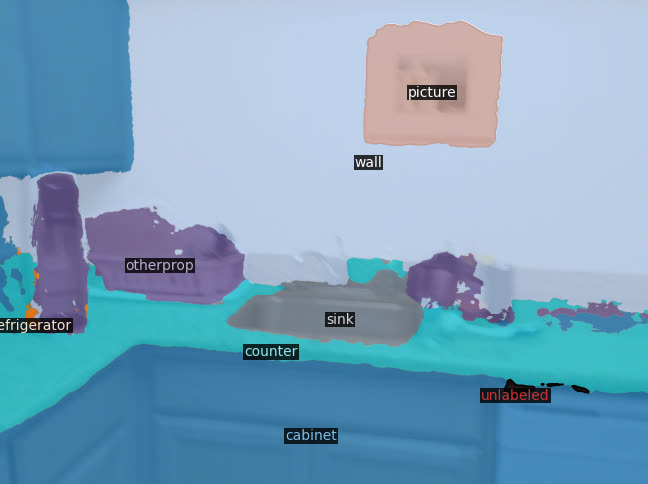} &
    \includegraphics[width=\lnw\linewidth]{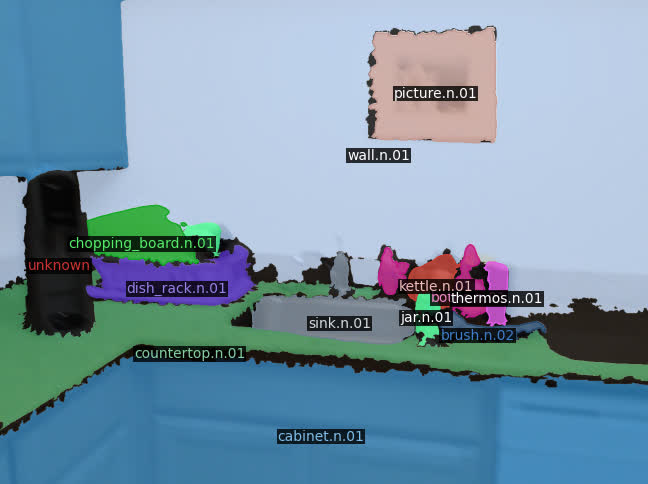} \\
    \includegraphics[width=\lnw\linewidth]{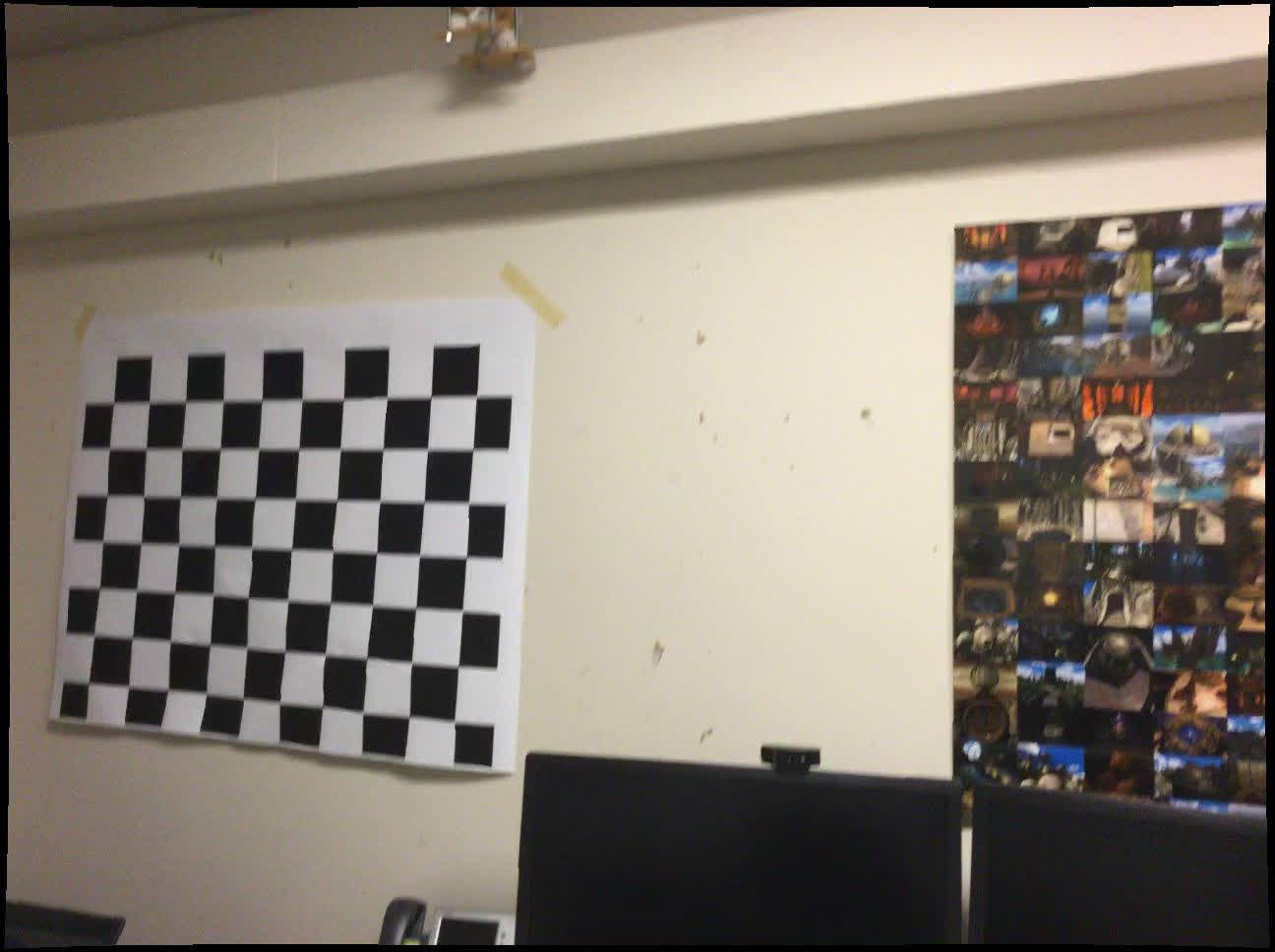} &
    \includegraphics[width=\lnw\linewidth]{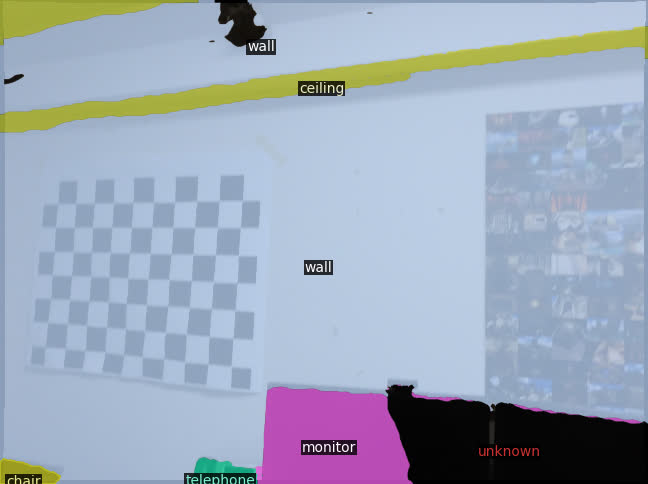} &
    \includegraphics[width=\lnw\linewidth]{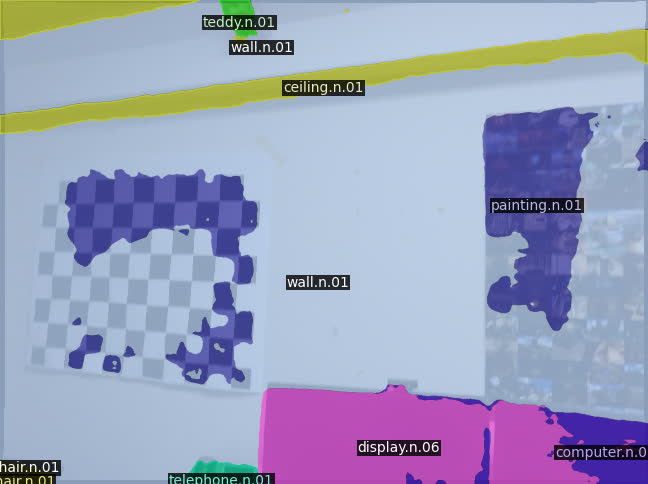} &
    \includegraphics[width=\lnw\linewidth]{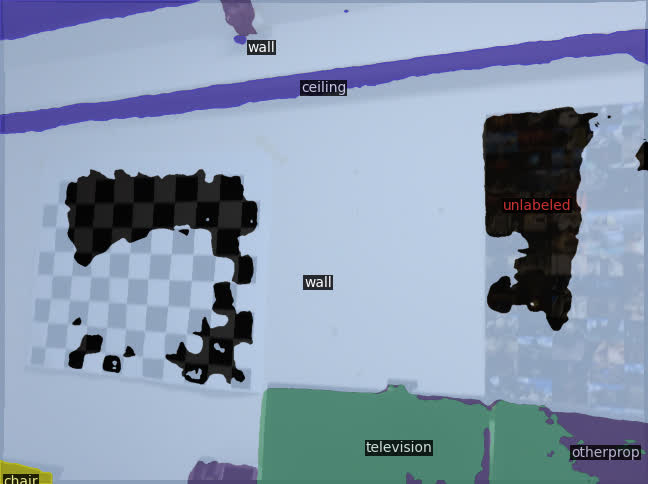} &
    \includegraphics[width=\lnw\linewidth]{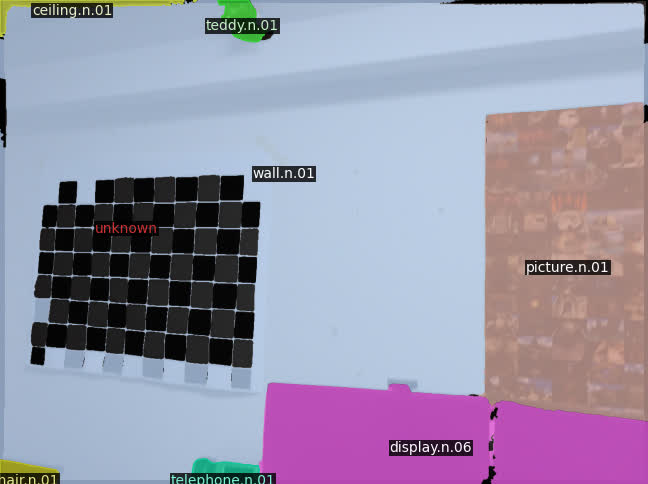}
\end{tabular}
\vspace{-5px}
\caption{\name{} generates more accurate and more complete labels compared to the labels annotated by humans and provided by ScanNet. 
Particularly, unlabeled sections in ScanNet are correctly filled in and many wrong annotations such as missing rogs and pictures are corrected. The output labels can then be projected into differnet label spaces, such as our wordnet space or the NYU40 categories.
}
\label{fig:arkitscenes_quali}
\end{figure*}

\subsection{Comparison to State-of-the-Art}
In Tab.~\ref{tab:main_table}, we compare \name\ to the state-of-the-art baselines ScanNet and SemanticNeRF. 
We report mean intersection-over-union (mIoU), mean accuracy (mAcc), as well as total accuracy (tAcc). 
We evaluate the methods in 2D by comparing the renderings or labeled frames with renderings from the ground-truth 3D mesh and in 3D by mapping the 2D renderings onto the corresponding vertices in the 3D ground-truth mesh.
Further, we measure the metrics over two different label sets.
The NYU40 label set~\cite{silberman_indoor_2012} consists of 40 semantic classes representing the common indoor classes in the short tail of the label distribution.
The wordnet label set consists of 186 classes, therefore measuring performance also over the long tail of the label distribution.

\begin{figure*}
\setlength{\tabcolsep}{1px}%
\newcommand{\lnw}{0.30}
\centering
\begin{tabular}{ccc}
    \textbf{Scannet} & \textbf{\name{}} (Ours) & \textbf{Groundtruth} \\
    \includegraphics[width=\lnw\linewidth]{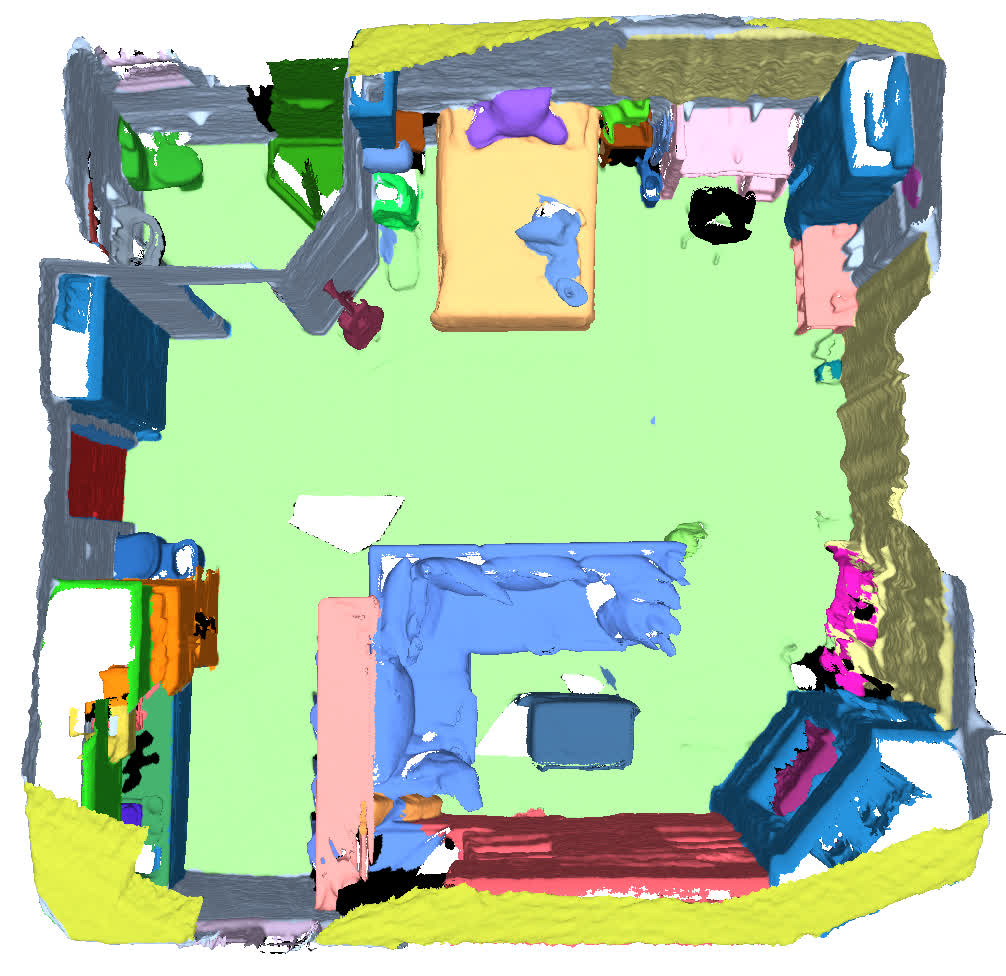} &
    \includegraphics[width=\lnw\linewidth]{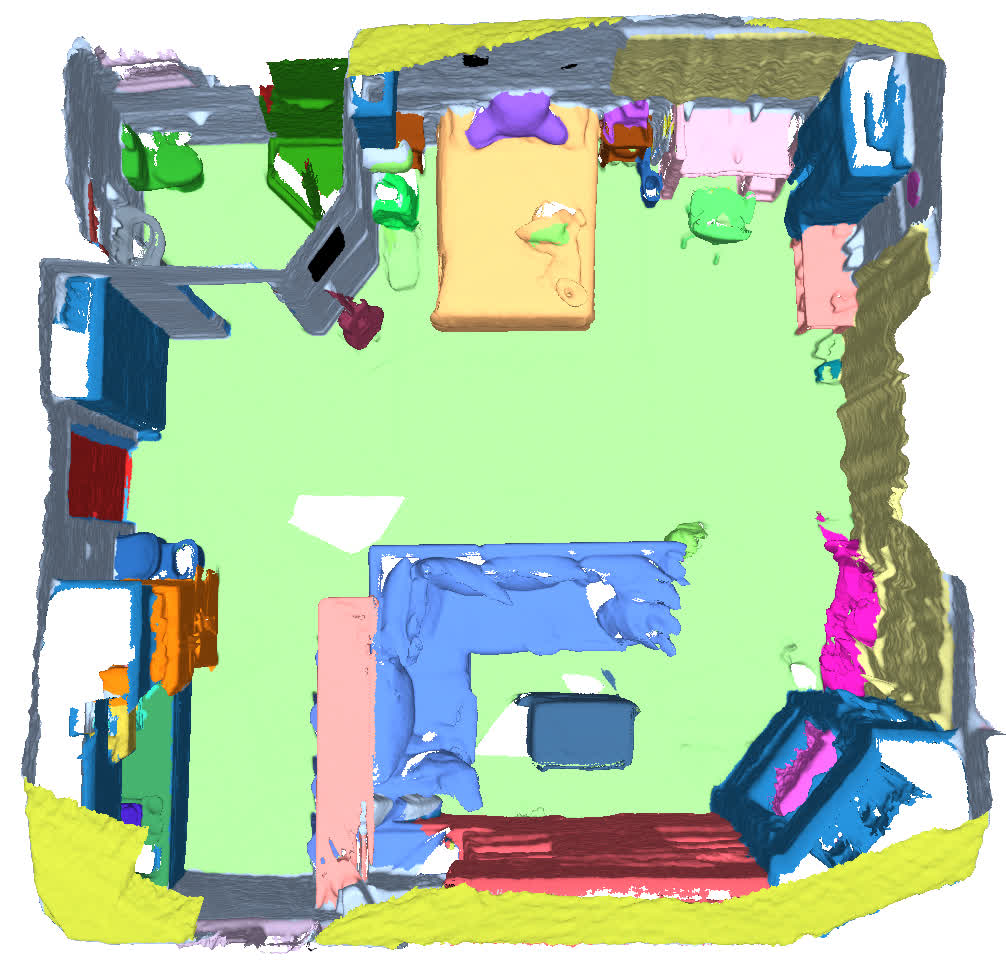} &
    \includegraphics[width=\lnw\linewidth]{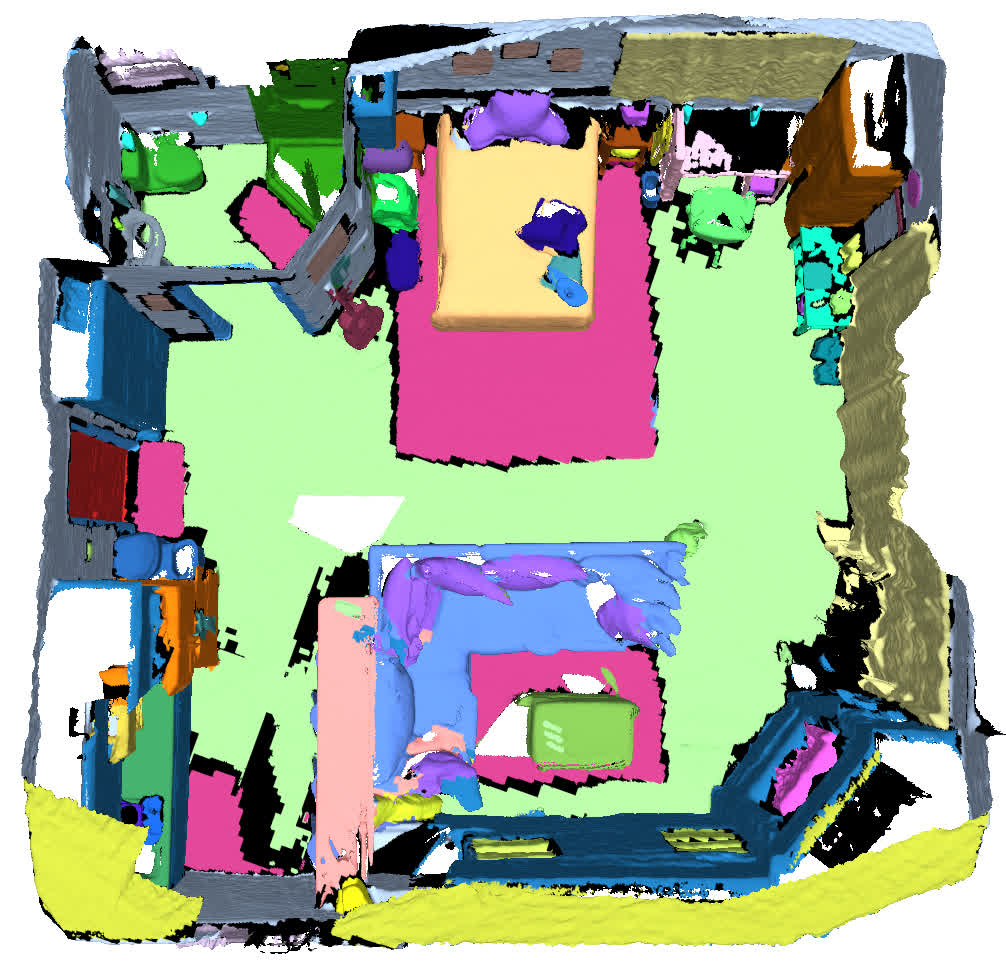} \\
    \vspace{-1mm}\includegraphics[width=\lnw\linewidth]{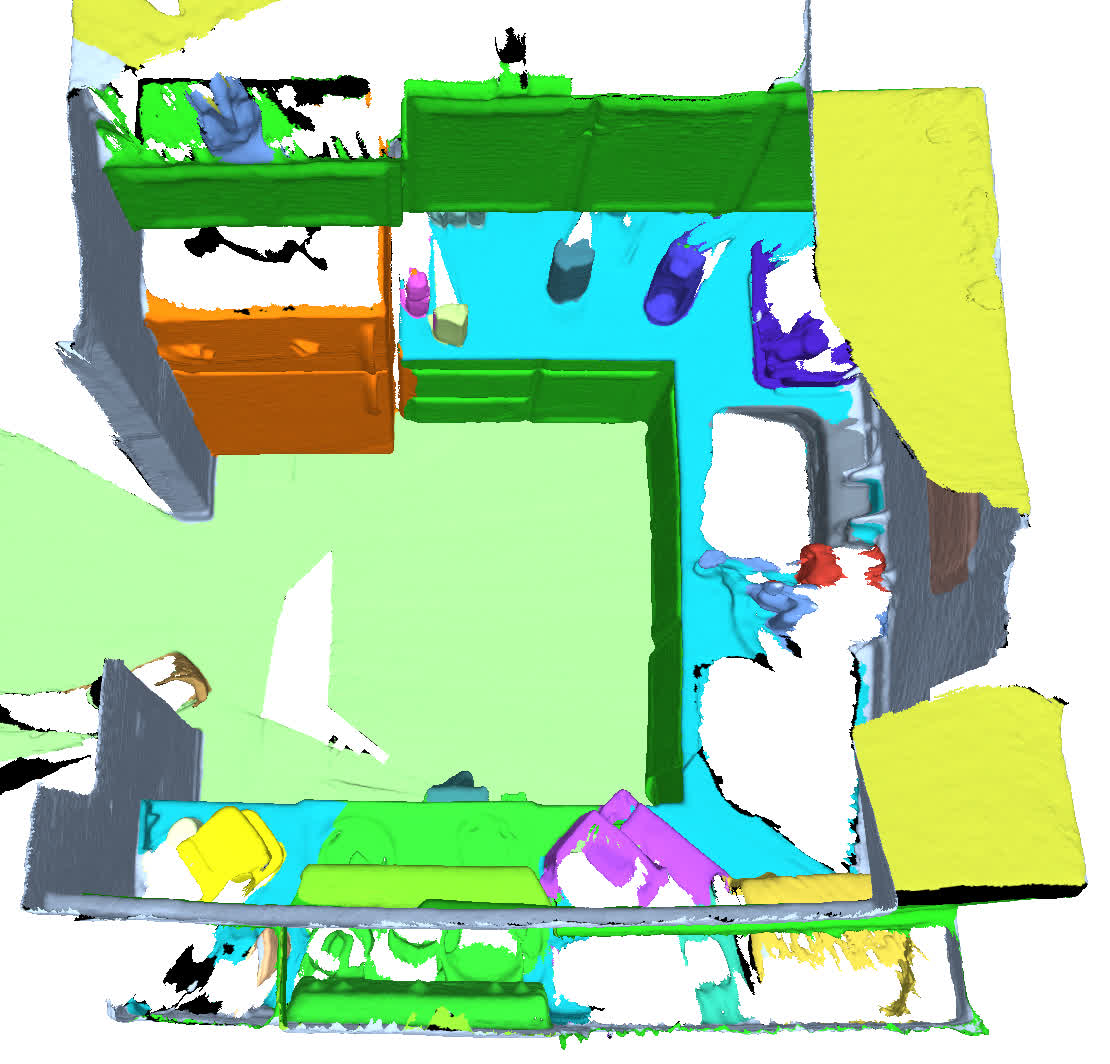} &
    \vspace{-1mm}\includegraphics[width=\lnw\linewidth]{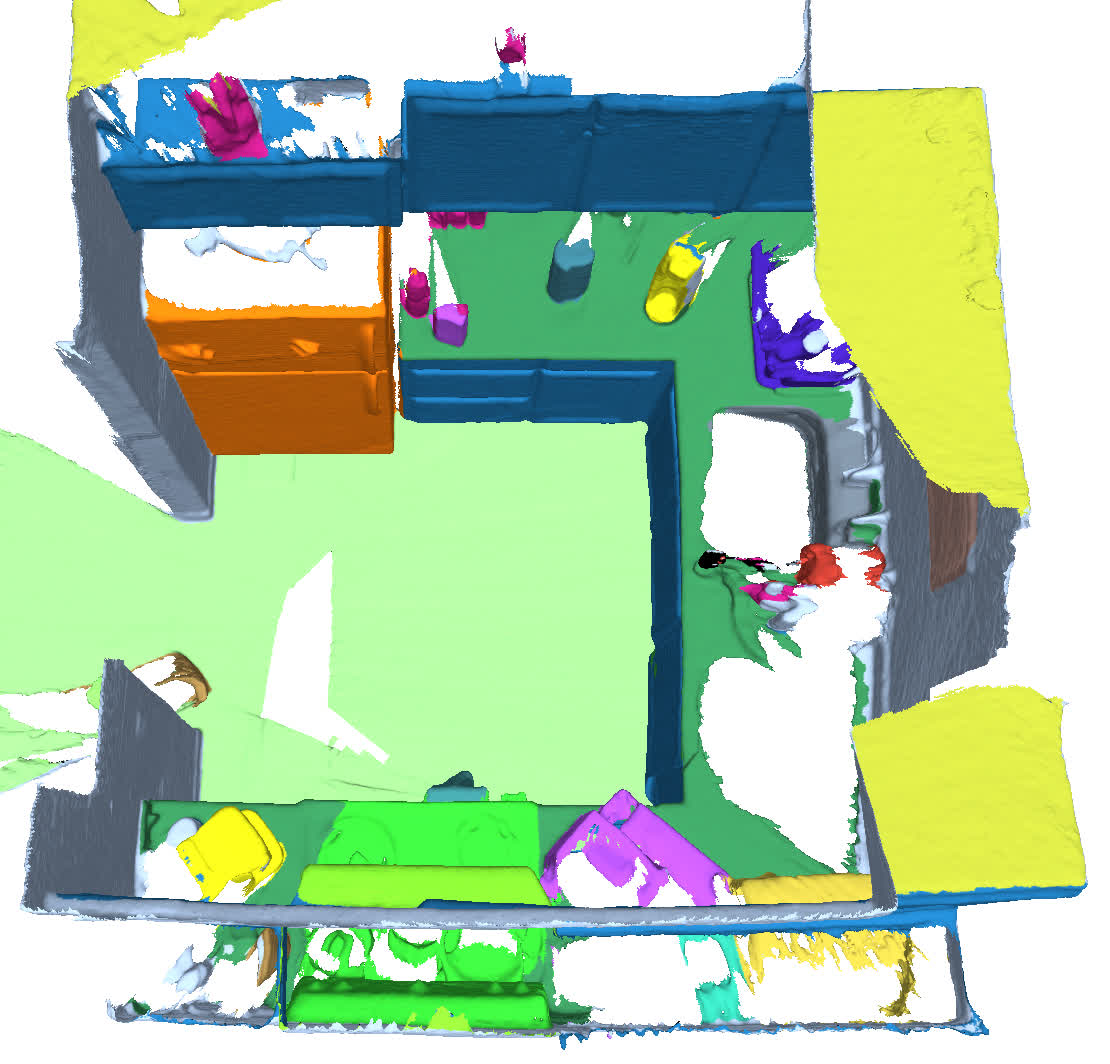} &
    \vspace{-1mm}\includegraphics[width=\lnw\linewidth]{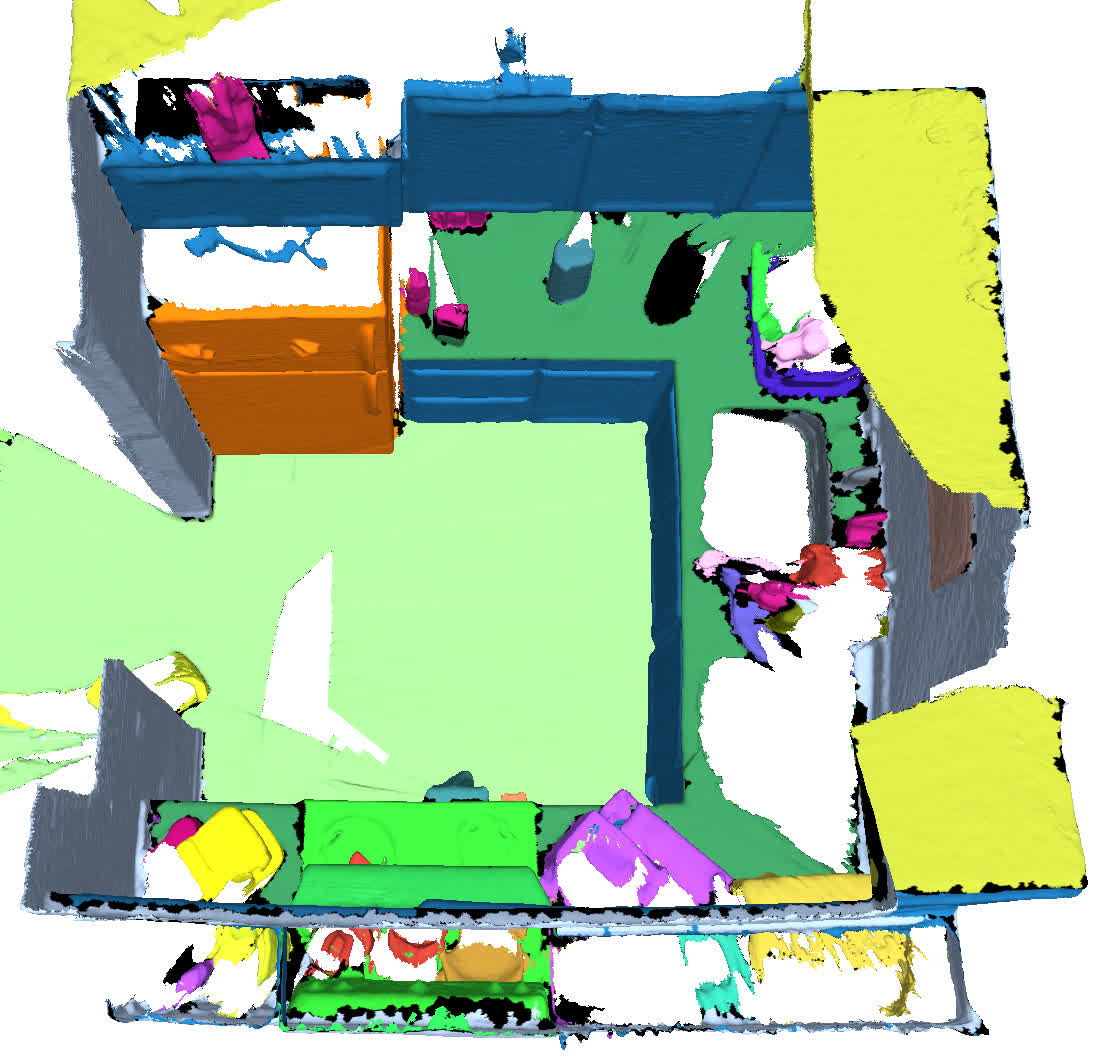} \\
    \includegraphics[width=\lnw\linewidth]{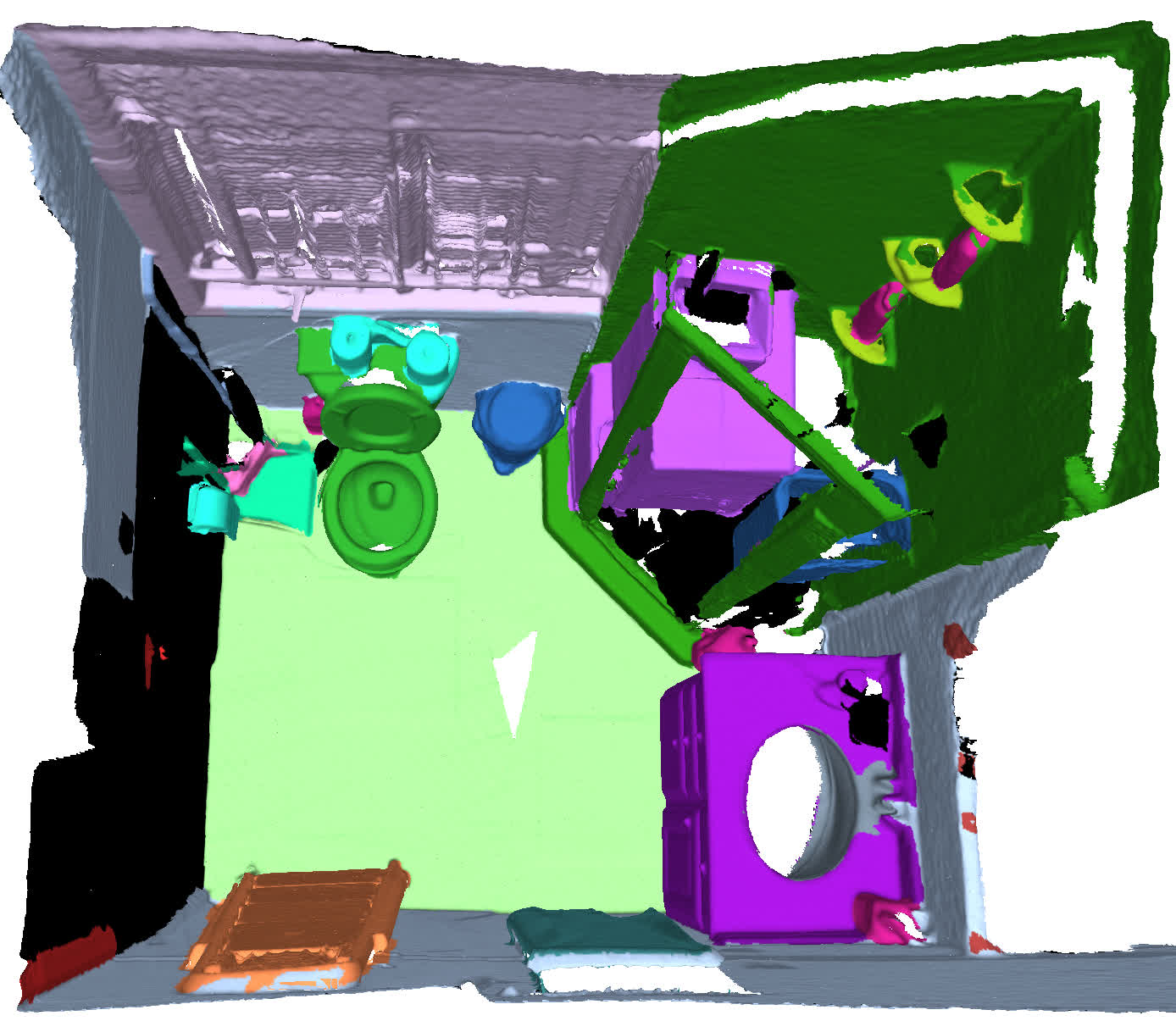} &
    \includegraphics[width=\lnw\linewidth]{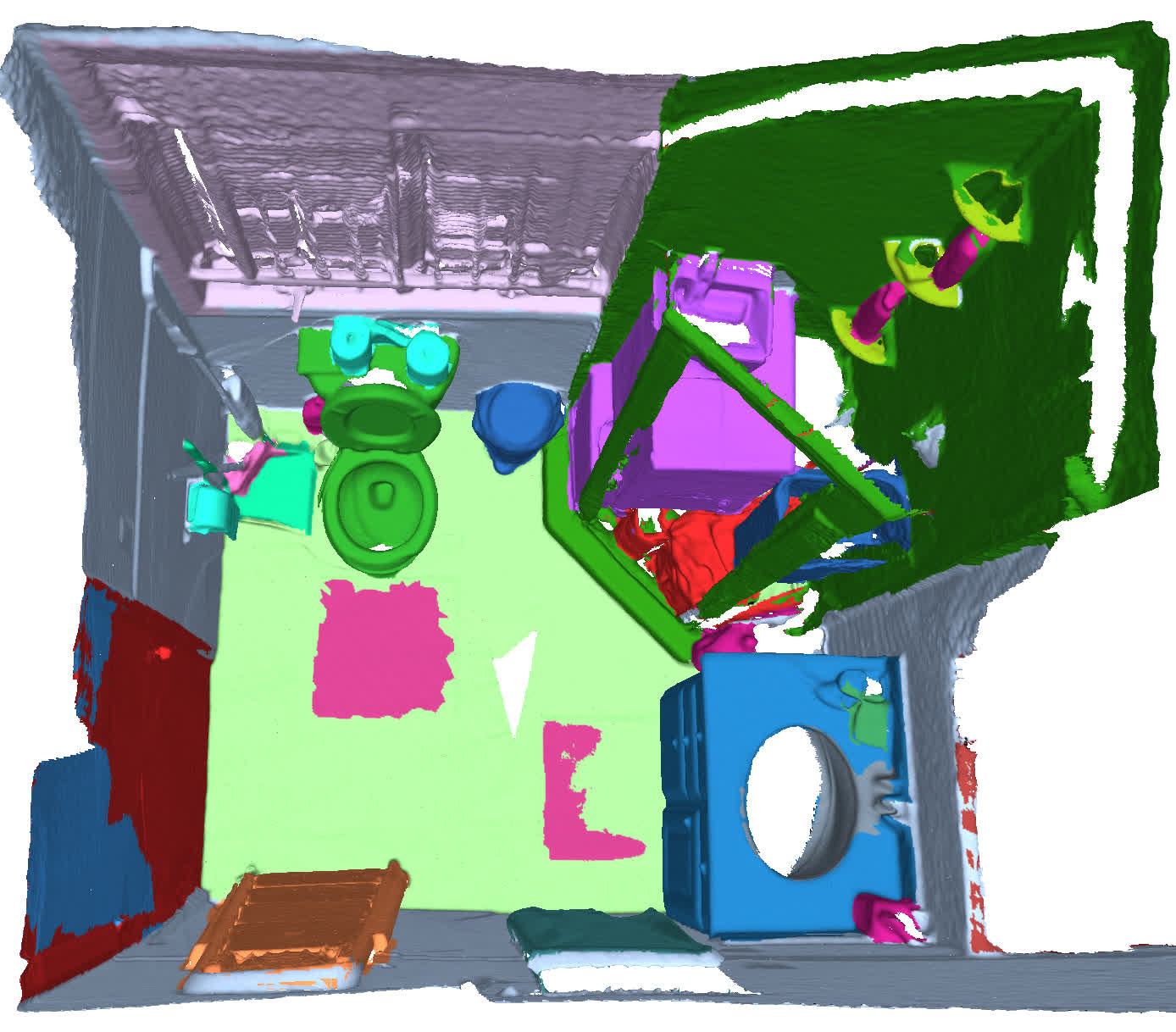} &
    \includegraphics[width=\lnw\linewidth]{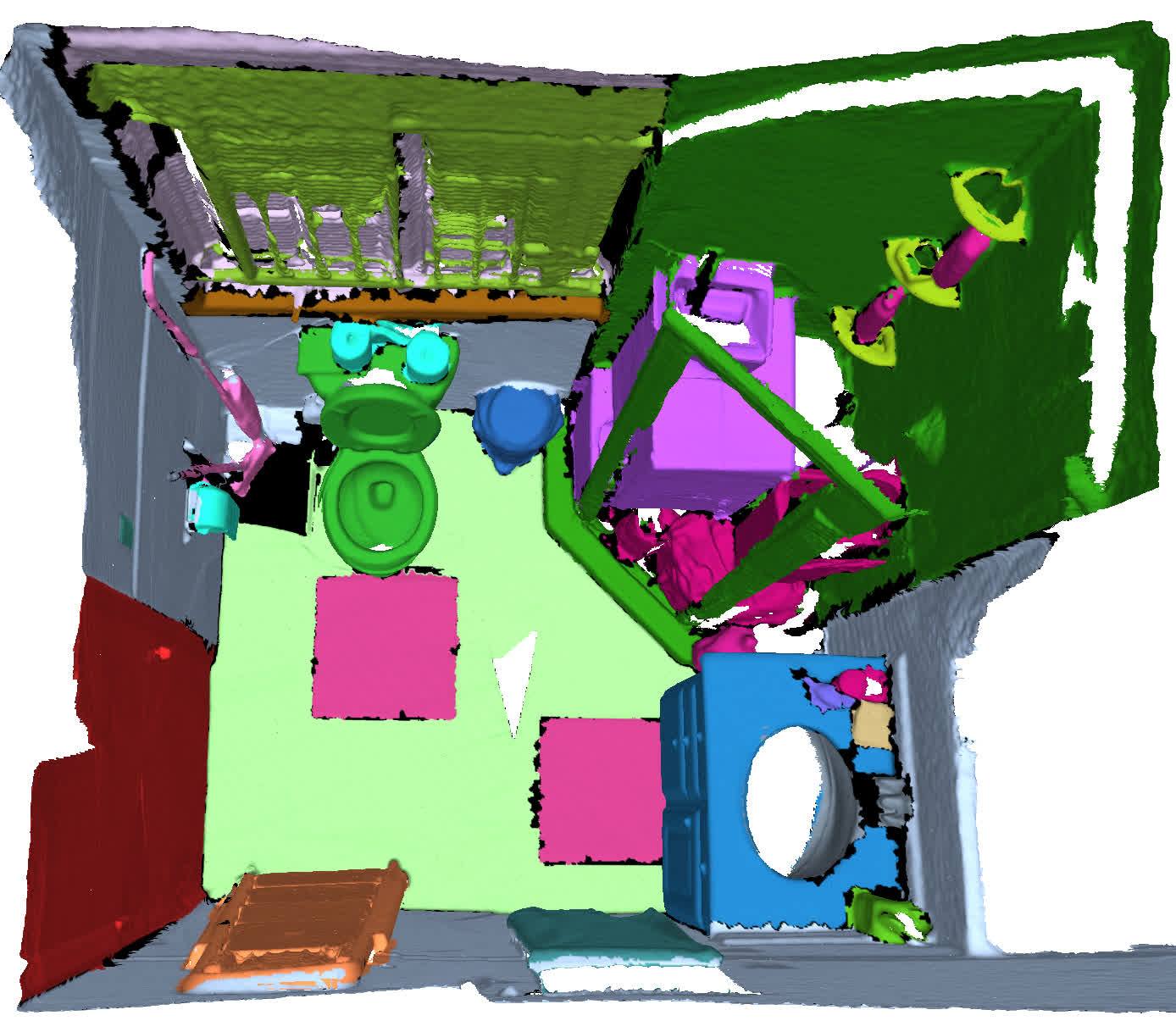} \\
\end{tabular}
\vspace{-5px}
\caption{\textbf{Dense 3D labels for ScanNetv2~\cite{scannet}.} 
We generate more consistent labels compared to human annotators and preserve rare classes (\eg{}, swivel chair in front of the desk).
Further, the labels are more complete (\eg{}, wall in bathroom)
and we can capture all object in the scene (\eg{}, dustpan in bathroom).
}
\label{fig:scannet_quali}
\end{figure*}
We show that our proposed pipeline generates better labels than human-annotated ScanNet labels and their lifted version through SemanticNeRF~\cite{zhi2021place}. 
Particularly, on the short tail of the distribution (NYU label set),
our pipeline significantly improves over the human annotated labels. 
This is due to more accurate object boundaries as well as more consistent and complete labels. 
For the long tail of the label distribution,
our method also outperforms all existing baselines
indicating that different 2D expert votes and 3D aggregation boosts
the quality of the annotated labels.   
Finally, we show that our fully automatic pipeline outperforms human annotations
on NYU40 classes, highlighting the potential of \name{} to generate labels at scale.

\parag{Qualitative comparison with ScanNet~\cite{scannet}}
In Fig.~\ref{fig:scannet_quali}, we compare qualitative results for ScanNet~\cite{scannet} with \name , and our groundtruth. 
To this end, we mapped the 2D renderings onto the high-resolution ground-truth mesh by projecting the mesh vertices into all labels using a visibility check. 
One can see that our pipeline produces consistently more complete and correct labels than the human annotations provided by ScanNet~\cite{scannet}.
\textit{E.g.}, our method consistently labels the kitchen countertop,
the mats in the bathroom,
and even the folded chair leaned against the desk.

\begin{figure*}
\setlength{\tabcolsep}{1px}%
\newcommand{\lnw}{0.25}
\centering
\begin{tabular}{cccc}
    \textbf{RGB} & \textbf{\name{}} 2D (Ours) & \textbf{Mask3D} & \textbf{\name{}} 3D (Ours) \\
    \includegraphics[width=\lnw\linewidth]{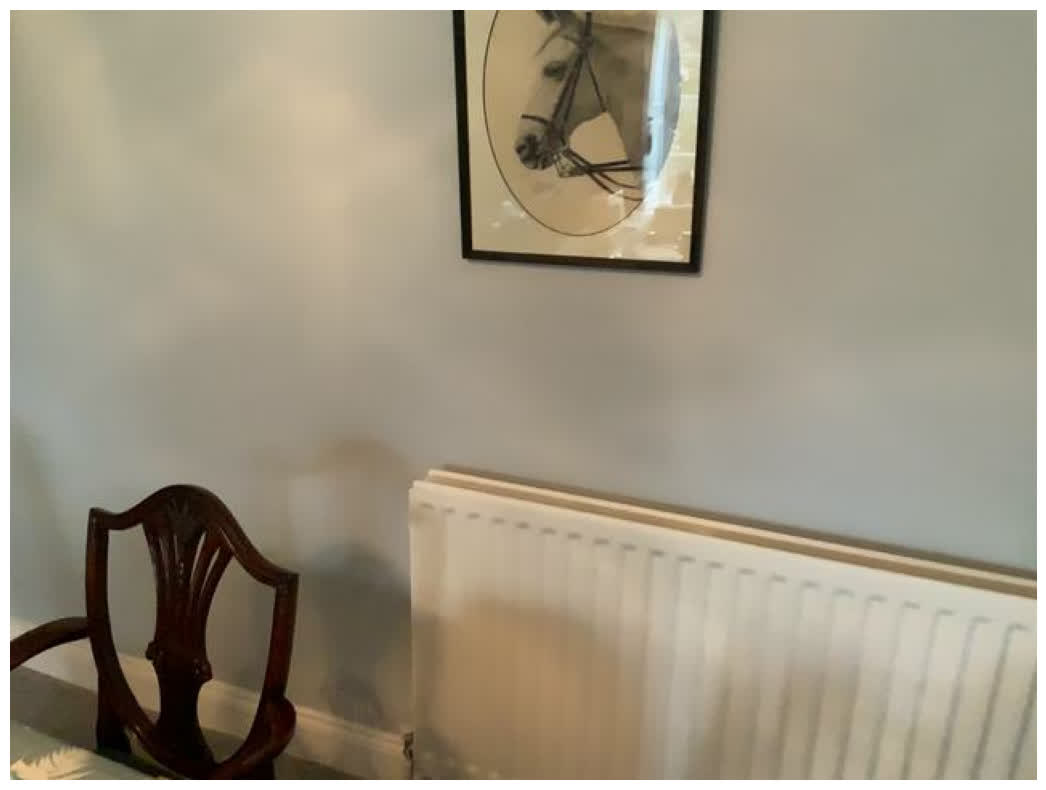}&%
    \includegraphics[width=\lnw\linewidth]{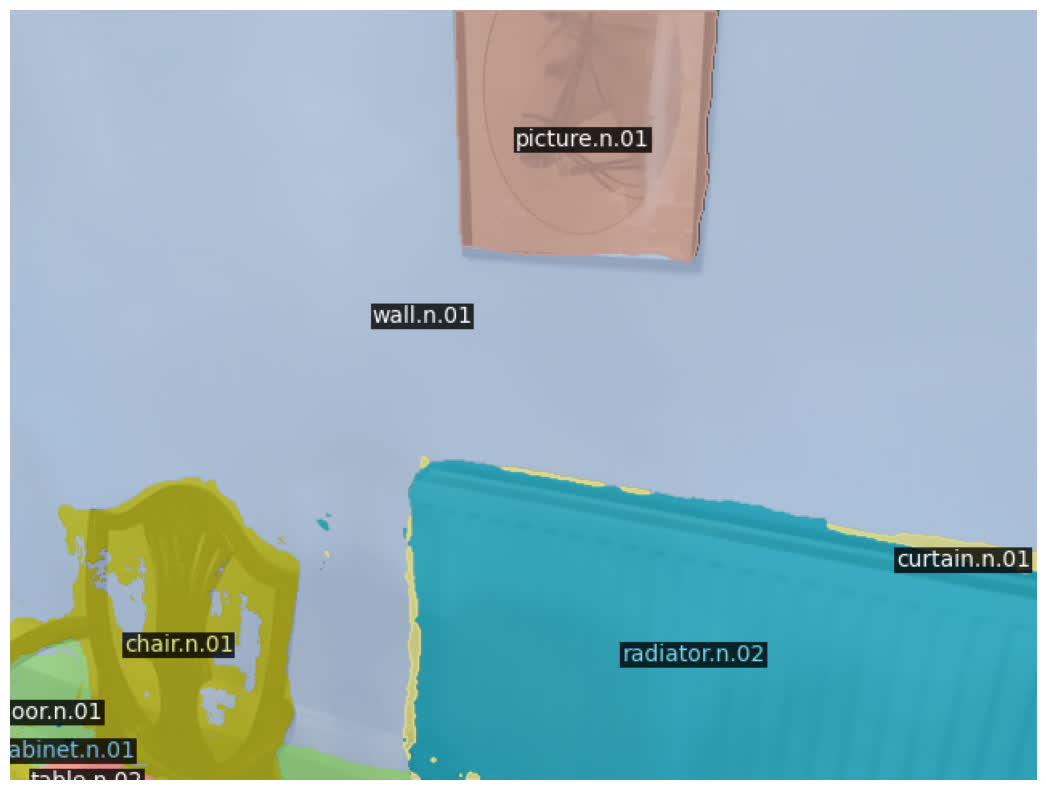}&%
    \includegraphics[width=\lnw\linewidth]{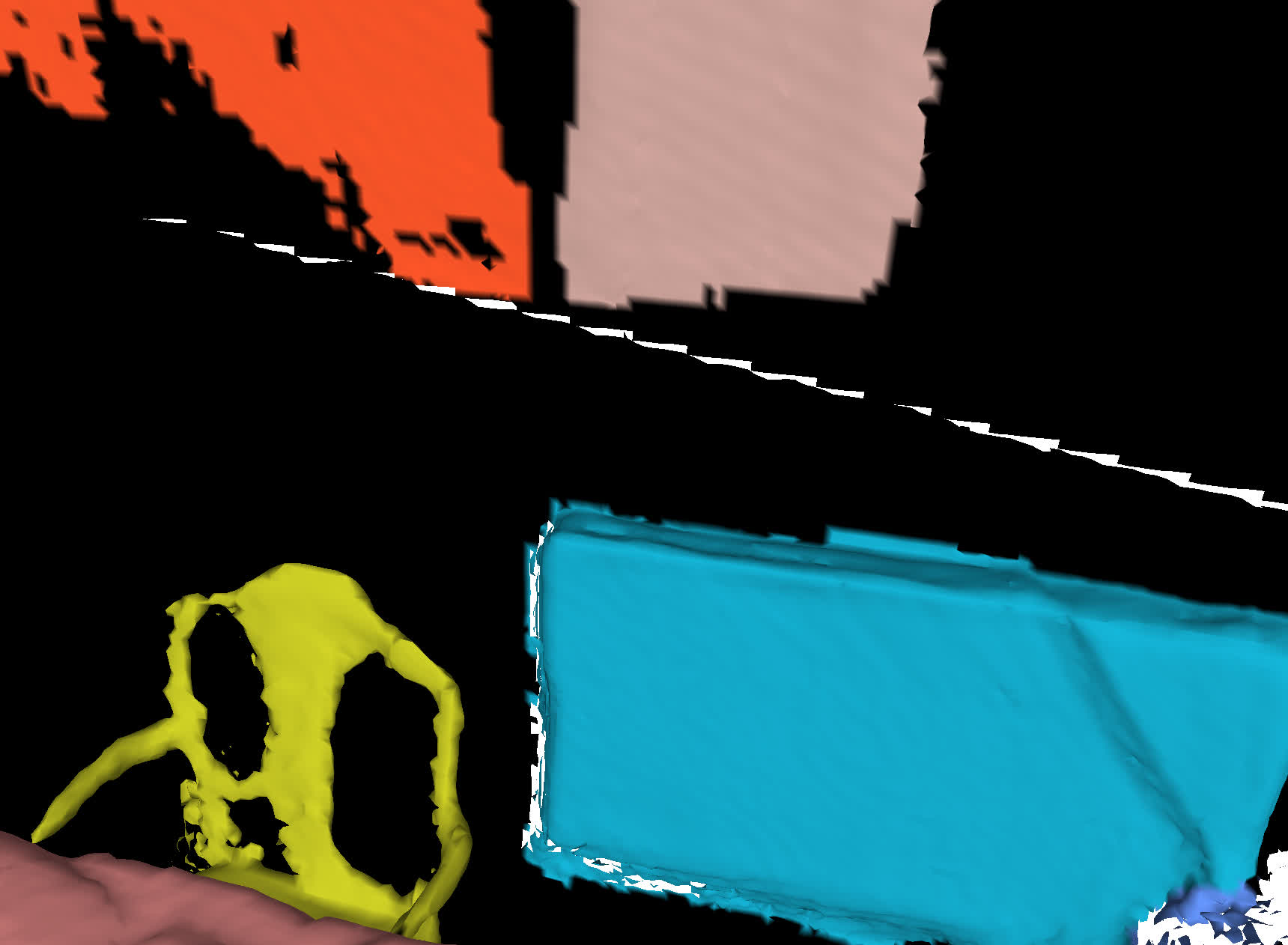}&%
    \includegraphics[width=\lnw\linewidth]{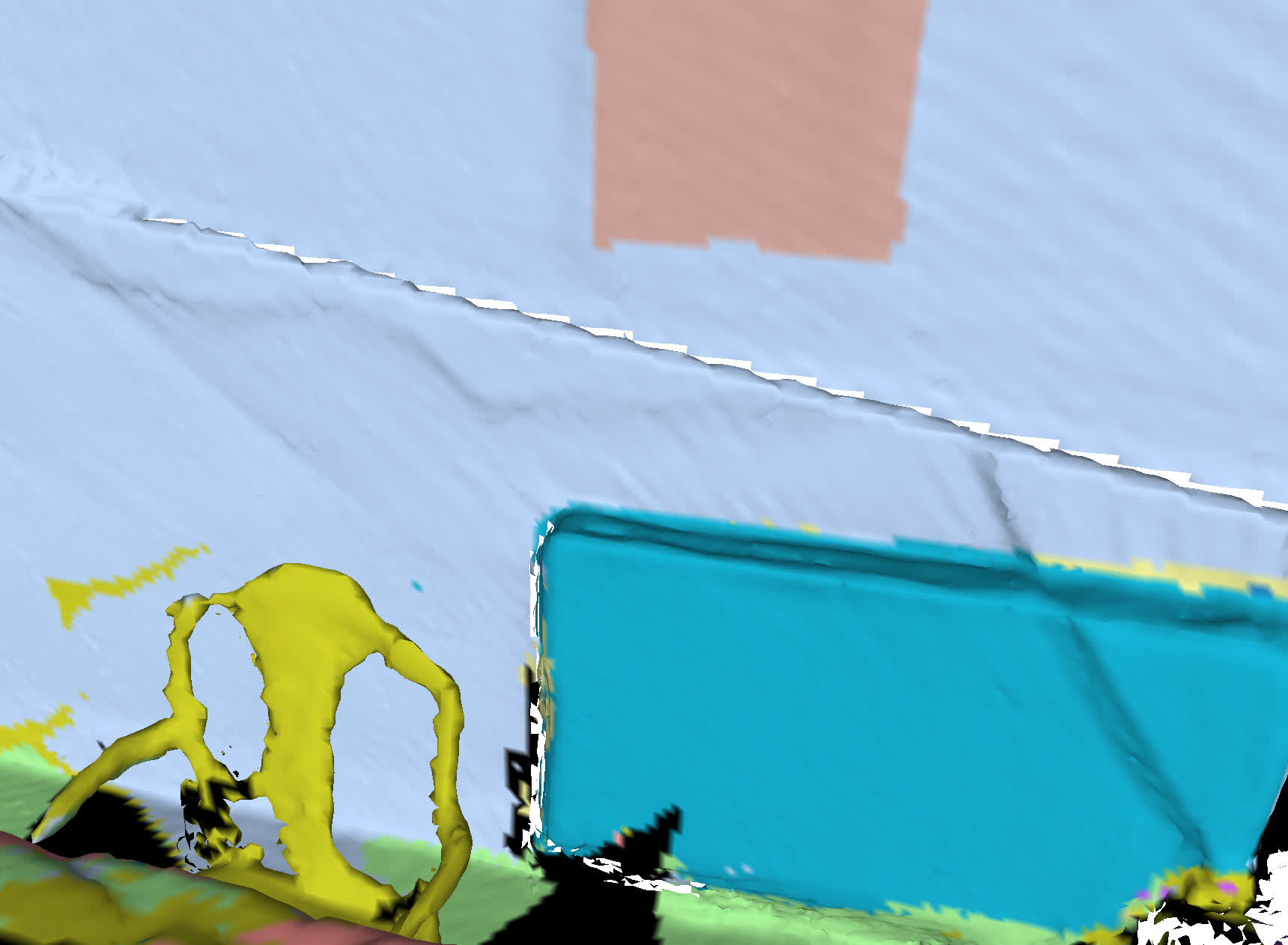}\\
    \includegraphics[width=\lnw\linewidth]{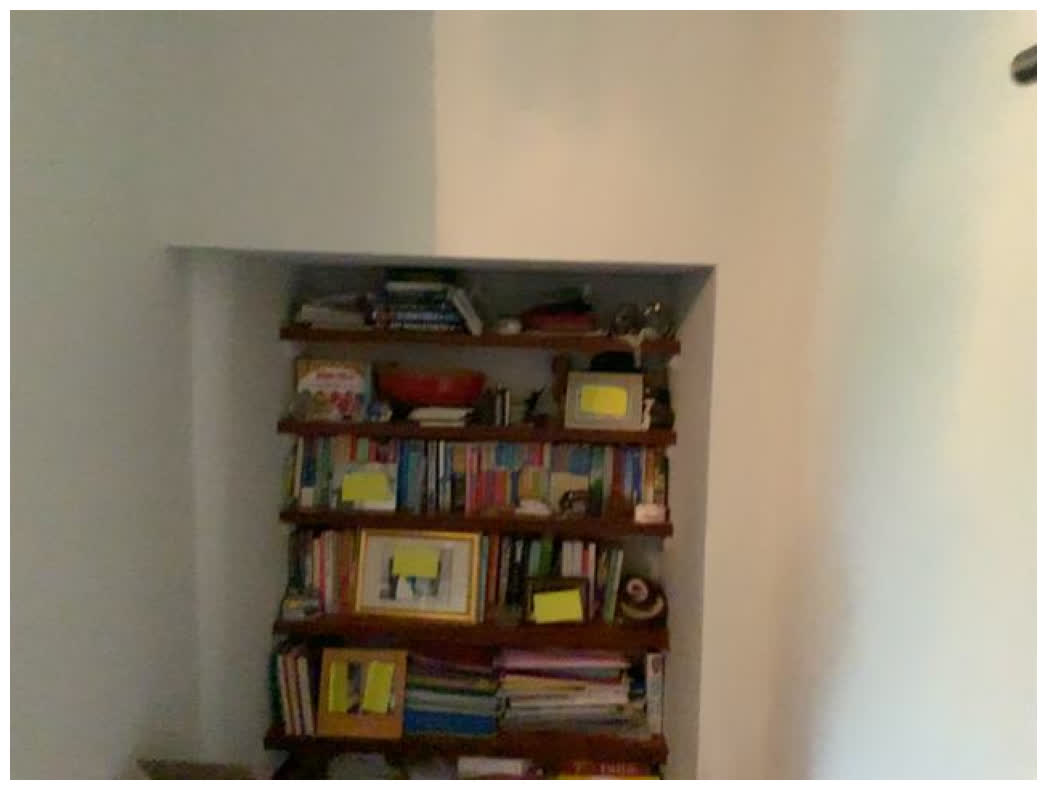}&
    \includegraphics[width=\lnw\linewidth]{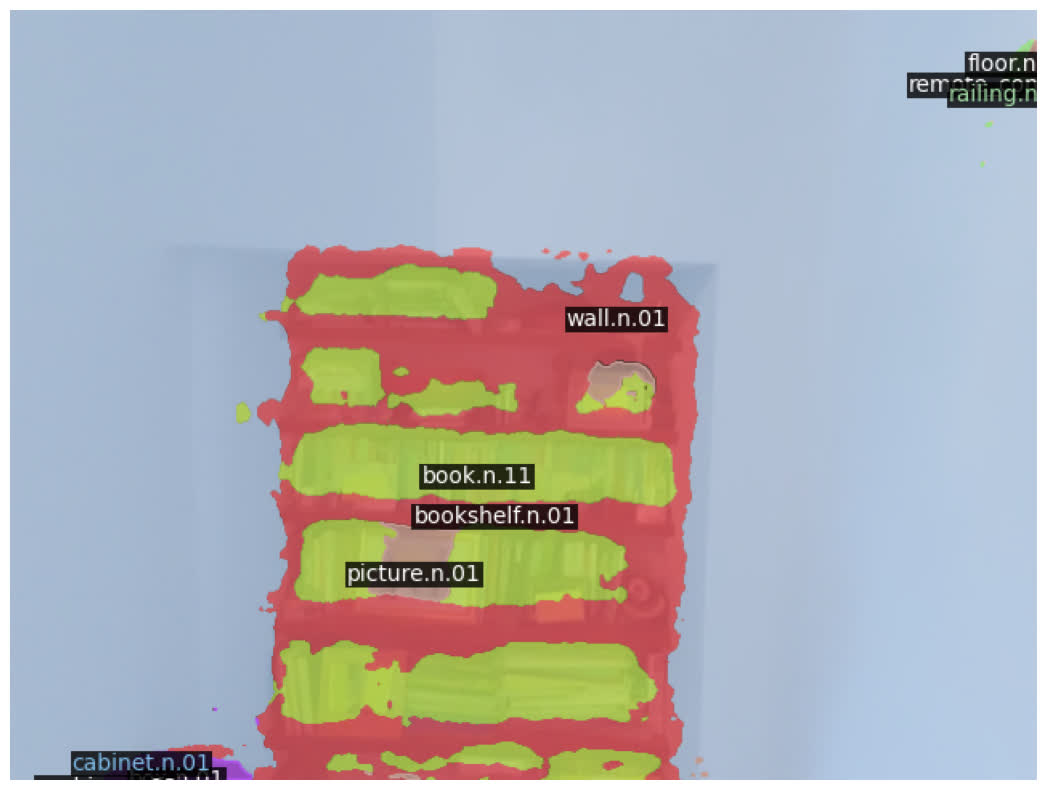}&
    \includegraphics[width=\lnw\linewidth]{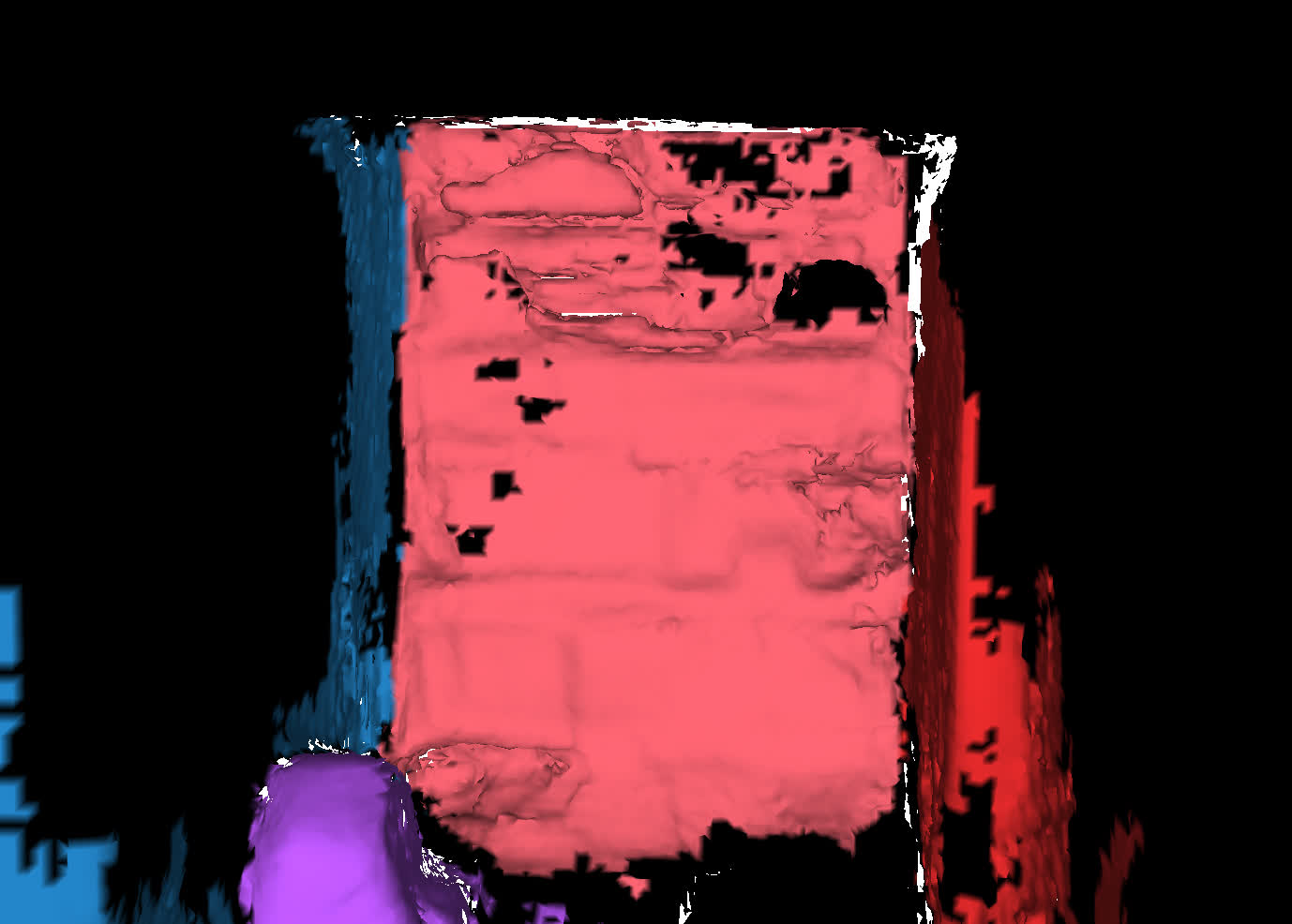}&
    \includegraphics[width=\lnw\linewidth]{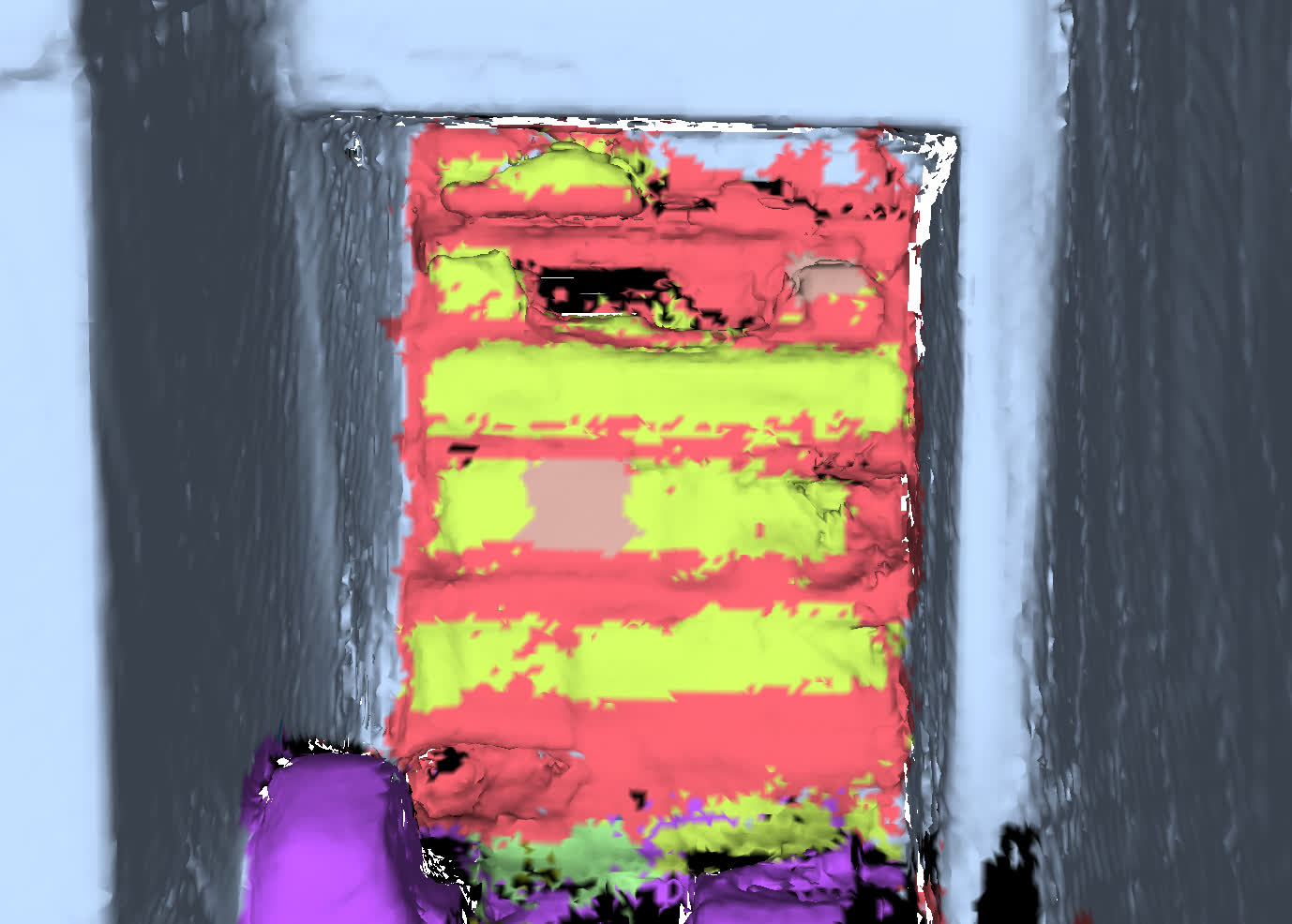}\\
    \includegraphics[width=\lnw\linewidth]{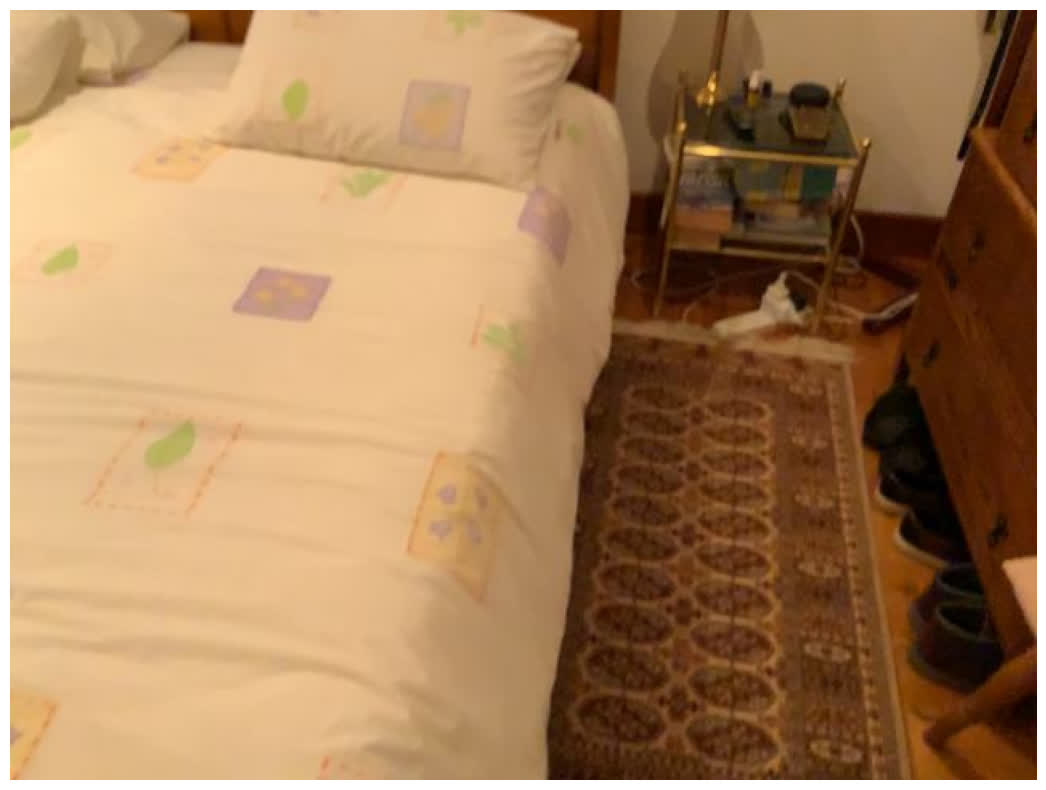}&
    \includegraphics[width=\lnw\linewidth]{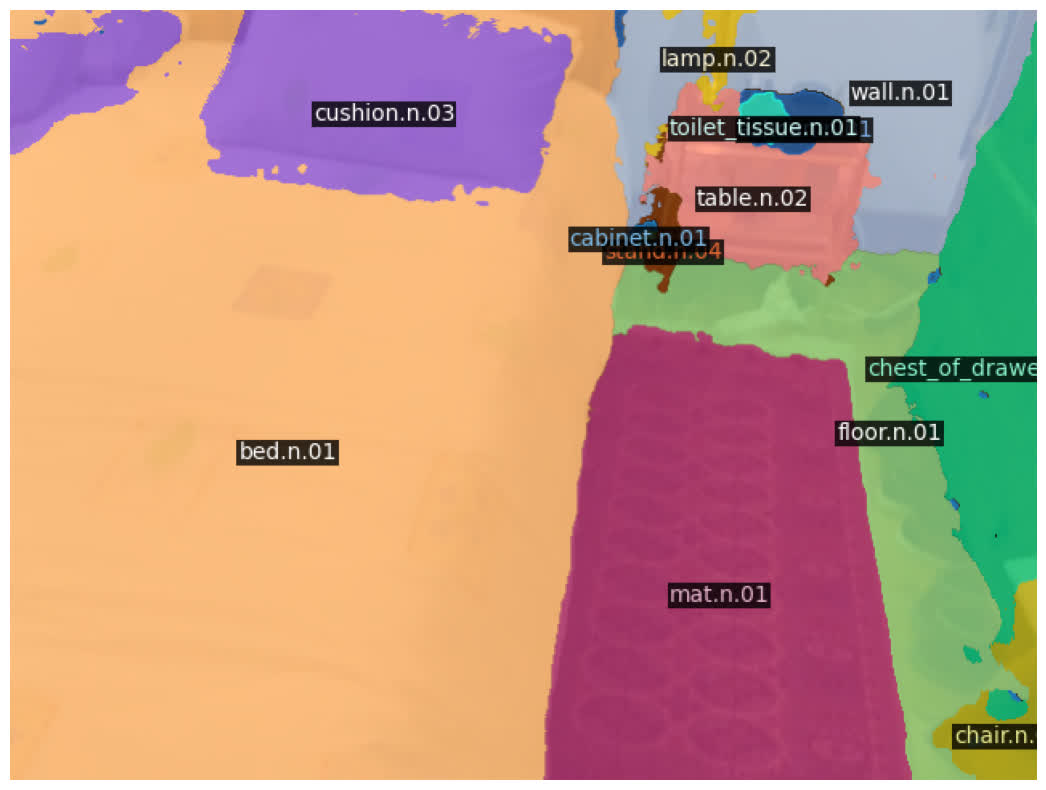}&
    \includegraphics[width=\lnw\linewidth]{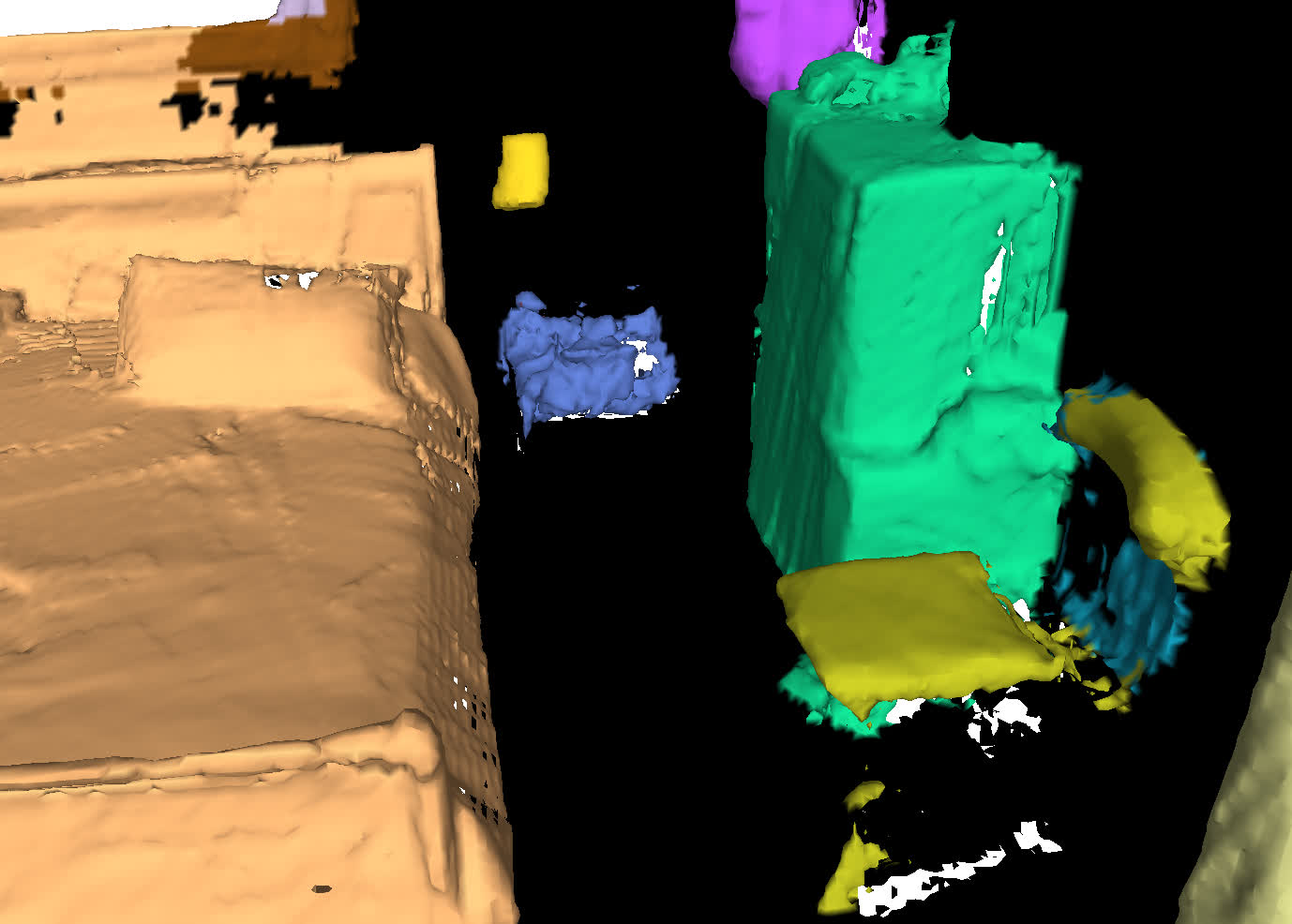}&
    \includegraphics[width=\lnw\linewidth]{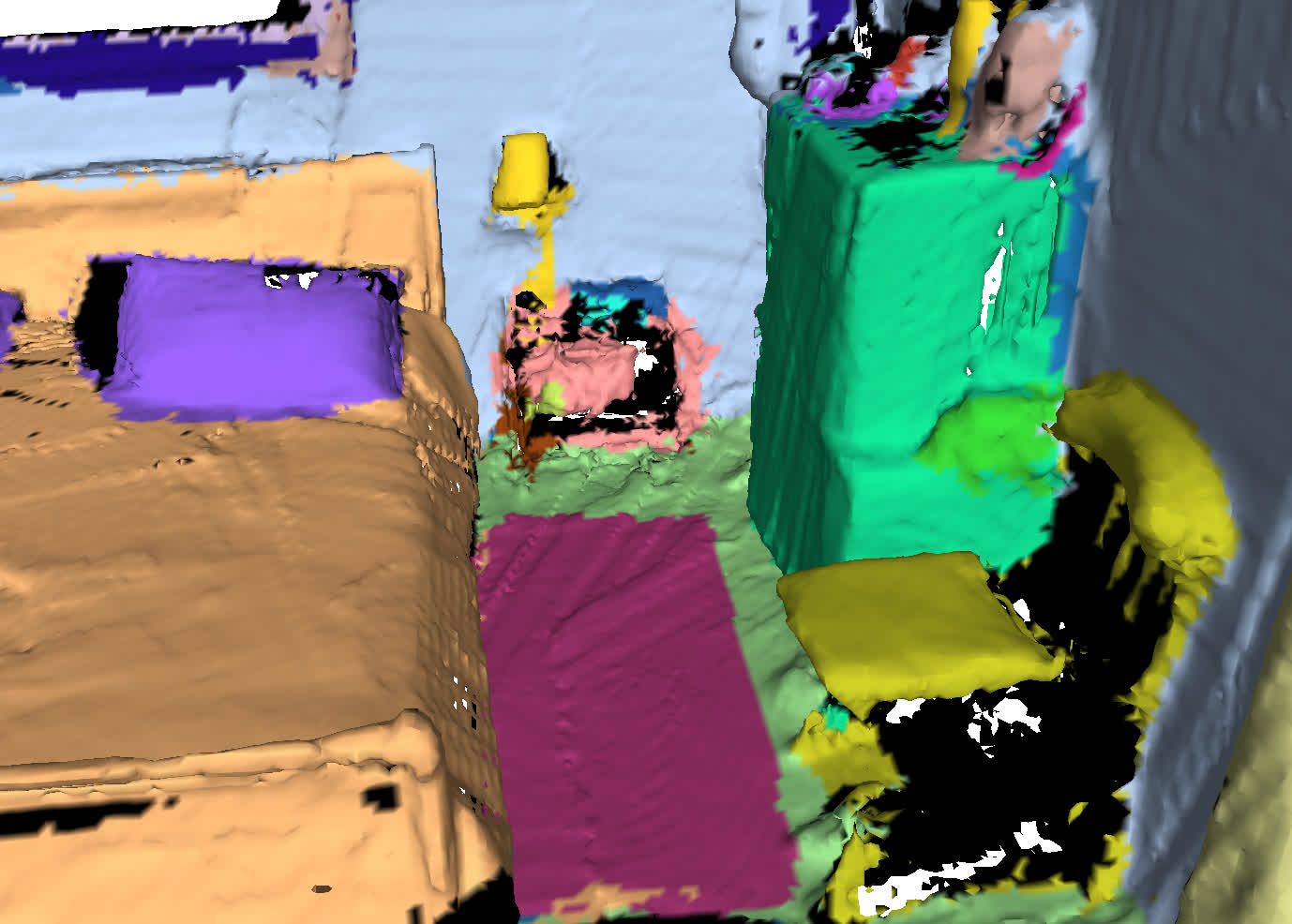}\\
\end{tabular}
\vspace{-6pt}
\caption{\textbf{Automatic dense labelling of ARKitScenes.}
We demonstrate the applicability to label RGB-D datasets that do not have dense labels available. 
Compared to state-of-the-art Mask3D~\cite{mask3d}, we generate dense annotations for all classes in the scene. 
Further, we segment on a higher level of detail (see picture and books in bookshelf, or objects on the cabinet/nightstand).
Thus, our labelling pipeline can readily be used on non-label dataset to provide training data for segmentation methods.
\vspace{-12px}}
\label{fig:arkitscenes_quali}
\end{figure*}
\begin{table}[]
\setlength{\tabcolsep}{1pt}
    \centering
    \small
    \begin{tabular}{lccc ccc}
    \toprule
    & \multicolumn{3}{c}{\textbf{ScanNet} \textit{(186 classes)}} & \multicolumn{3}{c}{\textbf{Replica} \textit{(150 classes)}}\\
        \cmidrule(r){2-4}  \cmidrule(r){5-7}
                    &  mIoU & mAcc & tAcc &  mIoU & mAcc & tAcc \\
                    \midrule
        OVSeg &  15.3 & 24.4 & 43.7 & 20.7 & 26.5 & 69.4 \\
        InternImage&  30.8 & 43.5 & 59.4 & 38.3 & 47.7 & 84.6 \\
        CMX &  28.2 & 41.0 & 54.2 & 17.0 & 38.0 & 84.6 \\
        Mask3D & 33.7 & 40.2 & 38.5 & 22.6 & 27.9 & 30.4 \\
        Consensus & 38.9 & 48.3 & 77.0 & 39.1 & 46.2 & 84.3 \\
        \name\ (ours) & \textbf{39.1} & \textbf{49.3} & \textbf{77.2} & \textbf{42.1} & \textbf{51.0} & \textbf{86.7} \\
        \bottomrule
    \end{tabular}  
    \vspace{-5px}
    \caption{Ablation of all base models in \name\ on our 5 labelled ScanNet~\cite{scannet} scenes and Replica~\cite{replica}. InternImage is the strongest single base model, but the fusion with other predictions and 3D lifting increases the accuracy considerably beyond any of the state-of-the-art single models.
    \vspace{-12px}}
    \label{tab:2d_model_performance}
\end{table}

\parag{ScanNet Label Quality}
Because our experiments require new high-accuracy annotations of ScanNet scenes,
we are able to estimate the quality of the default ScanNet labels.
As Tab.~\ref{tab:main_table} shows,
but also any human who inspects the ScanNet labels knows,
these are not perfect.
We argue in Sec.~\ref{subs:relabeling} that this reflects the open-world approach of the dataset and annotation workflow,
where -- exactly as in any real application -- semantics are ambiguous
and not always clearly defined.
We should also point out that even the detailed annotations we provide are not fully perfect.
However, given the background that the ScanNet labels are also used as a benchmark to compare accuracy of semantic classifiers,
our results indicate that a perfect prediction would reach accuracy values much lower than 100\%.
If two methods achieve higher mIoU on ScanNet than the ScanNet labels themselves,
it is not possible to draw a clear conclusion about which method is better.
This highlights the usefulness of improving the quality of the labels in datasets where some labels already exist.

\subsection{Ablation Study}

\parag{Does consensus voting make the model better?}
Tab.~\ref{tab:2d_model_performance} shows the evaluation on the standard metrics (mIoU, mAcc, tAcc) in 2D for the ScanNet and the Replica datasets.
We demonstrate that aggregating individual 2D predictions with our consensus voting mechanism improves upon the individual 2D models.
Further, we also show that lifting the 2D consensus into 3D using our optimization pipeline further improves the results compared to the individual 2D models.

\parag{Which model is the most important?}
Tab.~\ref{tab:2d_model_performance} shows that the performance of models differs noticeably.
Compared to the others, InternImage and Mask3D have the strongest positive impact on the segmentation quality. 
Additionally and unsurprisingly, Tab.~\ref{tab:main_table} shows that using ScanNet~\cite{scannet} labels as additional votes further improves performance.

\parag{Importance of 3D Lifting?}
We show in Tab.~\ref{tab:2d_model_performance} the effect of 3D lifting to aggregate semantic labels and make them multi-view consistent.
We compare \name{} with the aggregated consensus, as well as with individual models,and compute the 2D metrics on ScanNet and Replica.
One can see that the 3D lifting significantly improves the performance by at least $+1$ mIoU.

\subsection{Experiments on ARKitScenes}
To demonstrate the applicability of our labelling pipeline to new datasets, for which no dense labels exist, we run our pipeline on a set of scenes from the ARKitScenes~\cite{dehghan2021arkitscenes} dataset.
To this end, we process the smartphone trajectories using the low resolution depth maps as sensor depth and the corresponding VGA-resolution images as RGB input. 
We established these correspondences by synchronizing the depth and RGB timestamps.
In Fig.~\ref{fig:arkitscenes_quali}, we show qualitative results for 2 scenes of the data set.
One can see that the produced labels are more complete and accurate than for Mask3D, a state-of-the-art 3D instance segmentation method.
Thus, we demonstrate the feasibility of automatically labeling huge datasets with zero human intervention.

\section{Limitations}
\name{} is still limited to a fixed set of classes.
Extending it to output language embeddings instead of classes would make it more flexible and potentially help to resolve ambiguities.
The 3D lifting with SDFStudio has numerous hyper-parameters,
and this work possibly did not yet find the optimal settings.
In terms of accuracy, the pipeline can be further profit from newly developed models as research progresses, which will improve the output quality.
An interesting next step would be to implement a feedback loop where \name{} is used to produce a vast amount of automatically labeled training data, on which an additional model is trained as a distillation of the model zoo.

\section{Conclusion}
\vspace{-4pt}
We present a fully automatic labeling pipeline that generates semantic annotations of similar quality to human annotations,
with zero manual human labeling effort.
The method also improves the accuracy and consistency of existing annotations.
We quantitatively validate the performance of our pipeline on the ScanNet and Replica datasets.
On ScanNet, it outperforms the existing human annotations,
and on Replica it improves over all baseline methods.
Finally, we showcase the applicability to large-scale 3D datasets
and label images and point clouds of ARKitScenes.

\parag{Acknowledgments} This project is partially funded by the Swiss National Science Foundation (SNSF)
project TMAG-2\_216260, by two ETH Career Seed Awards ``ScanNetter" and ``Towards Open-World 3D Scene Understanding". 
FE is a postdoctoral research fellow at the ETH AI Center.

\newpage
{
\small
\bibliographystyle{ieeenat_fullname}
\bibliography{main}
}

\appendix
\section*{Appendix of `LabelMaker'}

The paper supplement consists of

\begin{itemize}
    \item anonymized source code
    \item video supplement with an additional explanaition of LabelMaker and full renderings of the LabelMaker output in all scenes
    \item additional experimental details in the following sections: \begin{itemize}
        \item the selected scenes from Replica, ScanNet and ARKitScenes,
        \item an explanation of our curated label mappings with some examples,
        \item per-category extension of our results on ScanNet,
        \item implementation details of the NeRF, 
        \item and our used annotation definitions.
    \end{itemize}
\end{itemize}

\section{Full qualitative examples}
We present full qualitative examples of all evaluated ScanNet trajectories in the video supplement.

\section{Code Supplement}

As part of the supplement, we also provide an anonymized version of the code base. It consists of a small library to match and evaluate different label spaces, and all code to run the \name{} pipeline in \texttt{scripts/}.

\section{Selected Scenes}
We use trajectories \texttt{scene0000\_00}, \texttt{scene0164\_02}, \texttt{scene0458\_00}, \texttt{scene0474\_01}, \texttt{scene0518\_00} from ScanNet and environments \texttt{room\_0}, \texttt{room\_1}, \texttt{room\_2} from Replica. In the main paper, we additionaly show qualitative results on ARKitScenes \texttt{42445991} and \texttt{42897688}.

\section{Label Mapping Examples}
In the following, we give a few examples of our curated label mapping that enables us to jointly use multiple models that are trained on different datasets (and label categories):

\parag{Simple Example 1:}
ScanNet category 1 is called `wall'.
It is mapped on NYU40 category `wall' (id 1),
ADE20k category `wall' (id 0),
Replica category `wall' (id 93),
and wordnet synkey \texttt{wall.n.01},
which we assign to our id 1.

\parag{Simple Example 2:}
ScanNet category 56 is called `trash can'.
It gets mapped on NYU40 category `otherfurniture' (id 39),
ADE20k category `ashcan' (id 138),
Replica category `bin' (id 10),
and wordnet synkey \texttt{ashcan.n.01} (synonyms ashcan, trash can, garbage can, wastebin, ash bin, ash-bin, ashbin, dustbin, trash barrel, trash bin), which we assign to our id 7.

\parag{Example of many-to-one mapping:}
In addition to the above examples, we also map \eg{}, ScanNet category `recycling bin' (id 97) to the wordnet synkey \texttt{ashcan.n.01} and the listed categories of the other datasets.

\parag{Example of many-to-many mapping}: We map ScanNet categories `pillow' (id 13), `couch cushions' (id 39), and `cushion' (id 39) all to wordnet synkey \texttt{cushion.n.03}. This gets mapped to NYU40 category `pillow' (id 18), ade classes `cushion' (id 39) and `pillow' (id 57), and Replica categories `cushion' (id 29) and `pillow' (id 61).

\section{Results for Individual Categories}

We present more detailed per-category data of our comparison to the ScanNet labels in Table~\ref{tab:scannet-nyu-perclass}.

\begin{table*}
\centering\small
\begin{tabular}{lrrrr}
\toprule
{} &  \name{} &  \name{} w/o ScanNet (automatic) &  SemanticNerf* &  ScanNet labels \\
\midrule
mIoU           &         0.534 &       0.507 &         0.452 &    0.477 \\
mAcc           &         0.650 &       0.640 &         0.566 &    0.562 \\
tAcc           &         0.775 &       0.753 &         0.693 &    0.692 \\
\midrule
wall           &         0.831 &       0.768 &         0.743 &    0.752 \\
floor          &         0.731 &       0.827 &         0.678 &    0.727 \\
cabinet        &         0.626 &       0.613 &         0.608 &    0.692 \\
bed            &         0.863 &       0.879 &         0.824 &    0.930 \\
chair          &         0.822 &       0.781 &         0.522 &    0.555 \\
sofa           &         0.797 &       0.807 &         0.790 &    0.811 \\
table          &         0.433 &       0.303 &         0.395 &    0.415 \\
door           &         0.782 &       0.671 &         0.658 &    0.520 \\
window         &         0.362 &       0.336 &         0.331 &    0.354 \\
bookshelf      &         0.235 &       0.442 &         0.500 &    0.528 \\
picture        &         0.510 &       0.024 &         0.252 &    0.255 \\
counter        &         0.682 &       0.524 &         0.640 &    0.639 \\
blinds         &         0.007 &       0.360 &         0.000 &    0.000 \\
desk           &         0.170 &       0.172 &         0.144 &    0.178 \\
curtain        &         0.884 &       0.809 &         0.871 &    0.884 \\
pillow         &         0.348 &       0.416 &         0.337 &    0.350 \\
floormat       &         0.072 &       0.541 &         0.000 &    0.000 \\
clothes        &         0.005 &       0.000 &         0.000 &    0.000 \\
ceiling        &         0.733 &       0.767 &         0.741 &    0.782 \\
books          &         0.023 &       0.292 &         0.000 &    0.000 \\
refrigerator   &         0.938 &       0.922 &         0.907 &    0.951 \\
television     &         0.713 &       0.486 &         0.171 &    0.168 \\
paper          &         0.000 &       0.061 &         0.000 &    0.000 \\
towel          &         0.648 &       0.580 &         0.504 &    0.554 \\
box            &         0.757 &       0.551 &         0.748 &    0.806 \\
nightstand     &         0.313 &       0.269 &         0.302 &    0.267 \\
toilet         &         0.911 &       0.845 &         0.795 &    0.895 \\
sink           &         0.777 &       0.802 &         0.569 &    0.715 \\
lamp           &         0.307 &       0.278 &         0.484 &    0.491 \\
bag            &         0.630 &       0.509 &         0.743 &    0.764 \\
otherstructure &         0.782 &       0.423 &         0.000 &    0.000 \\
otherfurniture &         0.474 &       0.321 &         0.344 &    0.373 \\
otherprop      &         0.460 &       0.344 &         0.309 &    0.374 \\
\bottomrule
\end{tabular}
\caption{Class-by-class evaluation on ScanNet in the NYU40 label space, on our annotated ScanNet scenes. Large gains of \name{} with respect to the ScanNet labels can be found, e.g., in chair, door, books, and television classes.}
\label{tab:scannet-nyu-perclass}
\end{table*}

\section{SDFStudio Semantic Head Details and Parameters}

We implement the semantic head as a small 4 layer MLP in parallel to the RGB head. While the RGB head takes as input the direction and the field feature at the rendered location, the semantic head is only dependent on the field feature to force the semantics to be the same from all viewing directions. To render semantics, we take a simple weighted sum over the output of the semantic head along the ray.

In the following command, we report the whole set of parameters we use to run our adapted SDFStudio models in all scenes:

{
\tiny
\begin{verbatim}
ns-train neus-facto \
  --pipeline.model.sdf-field.use-grid-feature True \
  --pipeline.model.sdf-field.hidden-dim 256 \
  --pipeline.model.sdf-field.num-layers 2 \
  --pipeline.model.sdf-field.num-layers-color 2 \
  --pipeline.model.sdf-field.semantic-num-layers 4 \
  --pipeline.model.sdf-field.use-appearance-embedding False \
  --pipeline.model.sdf-field.geometric-init True \
  --pipeline.model.sdf-field.inside-outside True  \
  --pipeline.model.sdf-field.bias 0.8 \
  --pipeline.model.sdf-field.beta-init 0.3 \
  --pipeline.model.sensor-depth-l1-loss-mult 0.3 \
  --pipeline.model.sensor-depth-sdf-loss-mult 0.3 \
  --pipeline.model.sensor-depth-freespace-loss-mult 0.3 \
  --pipeline.model.mono-normal-loss-mult 0.02 \
  --pipeline.model.mono-depth-loss-mult 0.000 \
  --pipeline.model.semantic-loss-mult 0.1 \
  --pipeline.model.semantic-ignore-label 0 \
  --trainer.max-num-iterations 20001 \
  --pipeline.datamanager.train-num-rays-per-batch 2048 \
  --pipeline.model.eikonal-loss-mult 0.1 \
  --pipeline.model.background-model none \
  sdfstudio-data \
  --include-sensor-depth True \
  --include-semantics True \
  --include-mono-prior True
\end{verbatim}
}

\section{WordNet Labels}
We use the following wordnet synkeys and definitions to annotate ScanNet scenes:

\begin{description}
\footnotesize
\setlength{\itemsep}{-\parskip}
\item[wall.n.01]	an architectural partition with a height and length greater than its thickness; used to divide or enclose an area or to support another structure
\item[chair.n.01]	a seat for one person, with a support for the back
\item[book.n.11]	a number of sheets (ticket or stamps etc.) bound together on one edge
\item[cabinet.n.01]	a piece of furniture resembling a cupboard with doors and shelves and drawers; for storage or display
\item[door.n.01]	a swinging or sliding barrier that will close the entrance to a room or building or vehicle
\item[floor.n.01]	also flooring; the inside lower horizontal surface (as of a room, hallway, tent, or other structure)
\item[ashcan.n.01]	also trash\_can, garbage\_can, wastebin, ash\_bin, ash-bin, ashbin, dustbin, trash\_barrel, trash\_bin; a bin that holds rubbish until it is collected
\item[table.n.02]	a piece of furniture having a smooth flat top that is usually supported by one or more vertical legs
\item[window.n.01]	a framework of wood or metal that contains a glass windowpane and is built into a wall or roof to admit light or air
\item[bookshelf.n.01]	a shelf on which to keep books
\item[display.n.06]	also video\_display; an electronic device that represents information in visual form
\item[cushion.n.03]	a soft bag filled with air or a mass of padding such as feathers or foam rubber etc.
\item[box.n.01]	a (usually rectangular) container; may have a lid
\item[picture.n.01]	also image, icon, ikon; a visual representation (of an object or scene or person or abstraction) produced on a surface
\item[ceiling.n.01]	the overhead upper surface of a covered space
\item[doorframe.n.01]	also doorcase; the frame that supports a door
\item[desk.n.01]	a piece of furniture with a writing surface and usually drawers or other compartments
\item[swivel\_chair.n.01]	a chair that swivels on its base
\item[towel.n.01]	a rectangular piece of absorbent cloth (or paper) for drying or wiping
\item[sofa.n.01]	also couch, lounge; an upholstered seat for more than one person
\item[sink.n.01]	plumbing fixture consisting of a water basin fixed to a wall or floor and having a drainpipe
\item[backpack.n.01]	also back\_pack, knapsack, packsack, rucksack, haversack; a bag carried by a strap on your back or shoulder
\item[lamp.n.02]	a piece of furniture holding one or more electric light bulbs
\item[chest\_of\_drawers.n.01]	also chest, bureau, dresser; furniture with drawers for keeping clothes
\item[apparel.n.01]	also wearing\_apparel, dress, clothes; clothing in general
\item[armchair.n.01]	chair with a support on each side for arms
\item[bed.n.01]	a piece of furniture that provides a place to sleep
\item[curtain.n.01]	also drape, drapery, mantle, pall; hanging cloth used as a blind (especially for a window)
\item[mirror.n.01]	polished surface that forms images by reflecting light
\item[plant.n.02]	also flora, plant\_life; (botany) a living organism lacking the power of locomotion
\item[radiator.n.02]	heater consisting of a series of pipes for circulating steam or hot water to heat rooms or buildings
\item[toilet\_tissue.n.01]	also toilet\_paper, bathroom\_tissue; a soft thin absorbent paper for use in toilets
\item[shoe.n.01]	footwear shaped to fit the foot (below the ankle) with a flexible upper of leather or plastic and a sole and heel of heavier material
\item[bag.n.01]	a flexible container with a single opening
\item[bottle.n.01]	a glass or plastic vessel used for storing drinks or other liquids; typically cylindrical without handles and with a narrow neck that can be plugged or capped
\item[countertop.n.01]	the top side of a counter
\item[coffee\_table.n.01]	also cocktail\_table; low table where magazines can be placed and coffee or cocktails are served
\item[toilet.n.02]	also can, commode, crapper, pot, potty, stool, throne; a plumbing fixture for defecation and urination
\item[computer\_keyboard.n.01]	also keypad; a keyboard that is a data input device for computers; arrangement of keys is modelled after the typewriter keyboard
\item[fridge.n.01]	also fridge; a refrigerator in which the coolant is pumped around by an electric motor
\item[stool.n.01]	a simple seat without a back or arms
\item[computer.n.01]	also computing\_machine, computing\_device, data\_processor, electronic\_computer, information\_processing\_system; a machine for performing calculations automatically
\item[mug.n.04]	with handle and usually cylindrical
\item[telephone.n.01]	also phone, telephone\_set; electronic equipment that converts sound into electrical signals that can be transmitted over distances and then converts received signals back into sounds
\item[light.n.02]	also light\_source; any device serving as a source of illumination
\item[jacket.n.01]	a short coat
\item[bathtub.n.01]	also bathing\_tub, bath, tub; a relatively large open container that you fill with water and use to wash the body
\item[shower\_curtain.n.01]	a curtain that keeps water from splashing out of the shower area
\item[microwave.n.02]	also microwave\_oven; kitchen appliance that cooks food by passing an electromagnetic wave through it; heat results from the absorption of energy by the water molecules in the food
\item[footstool.n.01]	also footrest, ottoman, tuffet; a low seat or a stool to rest the feet of a seated person
\item[baggage.n.01]	also luggage; cases used to carry belongings when traveling
\item[laptop.n.01]	also laptop\_computer; a portable computer small enough to use in your lap
\item[printer.n.03]	also printing\_machine; a machine that prints
\item[shower\_stall.n.01]	also shower\_bath; booth for washing yourself, usually in a bathroom
\item[soap\_dispenser.n.01]	dispenser of liquid soap
\item[stove.n.01]	also kitchen\_stove, range, kitchen\_range, cooking\_stove; a kitchen appliance used for cooking food
\item[fan.n.01]	a device for creating a current of air by movement of a surface or surfaces
\item[paper.n.01]	a material made of cellulose pulp derived mainly from wood or rags or certain grasses
\item[stand.n.04]	a small table for holding articles of various kinds
\item[bench.n.01]	a long seat for more than one person
\item[wardrobe.n.01]	also closet, press; a tall piece of furniture that provides storage space for clothes; has a door and rails or hooks for hanging clothes
\item[blanket.n.01]	also cover; bedding that keeps a person warm in bed
\item[booth.n.02]	also cubicle, stall, kiosk; small area set off by walls for special use
\item[duplicator.n.01]	also copier; apparatus that makes copies of typed, written or drawn material
\item[bar.n.03]	a rigid piece of metal or wood; usually used as a fastening or obstruction or weapon
\item[soap\_dish.n.01]	a bathroom or kitchen fixture for holding a bar of soap
\item[switch.n.01]	also electric\_switch, electrical\_switch; control consisting of a mechanical or electrical or electronic device for making or breaking or changing the connections in a circuit
\item[coffee\_maker.n.01]	a kitchen appliance for brewing coffee automatically
\item[decoration.n.01]	also ornament, ornamentation; something used to beautify
\item[range\_hood.n.01]	exhaust hood over a kitchen range
\item[blackboard.n.01]	also chalkboard; sheet of slate; for writing with chalk
\item[clock.n.01]	a timepiece that shows the time of day
\item[railing.n.01]	also rail; a barrier consisting of a horizontal bar and supports
\item[mat.n.01]	-- merged with rug.n.01 -- a thick flat pad used as a floor covering
\item[seat.n.03]	furniture that is designed for sitting on
\item[bannister.n.02]	also banister, balustrade, balusters, handrail; a railing at the side of a staircase or balcony to prevent people from falling
\item[container.n.01]	any object that can be used to hold things (especially a large metal boxlike object of standardized dimensions that can be loaded from one form of transport to another)
\item[mouse.n.04]	also computer\_mouse; a hand-operated electronic device that controls the coordinates of a cursor on your computer screen as you move it around on a pad; on the bottom of the device is a ball that rolls on the surface of the pad
\item[person.n.02]	a human body (usually including the clothing)
\item[stairway.n.01]	also staircase; a way of access (upward and downward) consisting of a set of steps
\item[basket.n.01]	also handbasket; a container that is usually woven and has handles
\item[dumbbell.n.01]	an exercising weight; two spheres connected by a short bar that serves as a handle
\item[column.n.07]	also pillar; (architecture) a tall vertical cylindrical structure standing upright and used to support a structure
\item[bucket.n.01]	also pail; a roughly cylindrical vessel that is open at the top
\item[windowsill.n.01]	the sill of a window; the horizontal member at the bottom of the window frame
\item[signboard.n.01]	also sign; structure displaying a board on which advertisements can be posted
\item[dishwasher.n.01]	also dish\_washer, dishwashing\_machine; a machine for washing dishes
\item[loudspeaker.n.01]	also speaker, speaker\_unit, loudspeaker\_system, speaker\_system; electro-acoustic transducer that converts electrical signals into sounds loud enough to be heard at a distance
\item[washer.n.03]	also automatic\_washer, washing\_machine; a home appliance for washing clothes and linens automatically
\item[paper\_towel.n.01]	a disposable towel made of absorbent paper
\item[clothes\_hamper.n.01]	also laundry\_basket, clothes\_basket, voider; a hamper that holds dirty clothes to be washed or wet clothes to be dried
\item[piano.n.01]	also pianoforte, forte-piano; a keyboard instrument that is played by depressing keys that cause hammers to strike tuned strings and produce sounds
\item[sack.n.01]	also poke, paper\_bag, carrier\_bag; a bag made of paper or plastic for holding customer's purchases
\item[handcart.n.01]	also pushcart, cart, go-cart; wheeled vehicle that can be pushed by a person; may have one or two or four wheels
\item[blind.n.03]	also screen; a protective covering that keeps things out or hinders sight
\item[dish\_rack.n.01]	a rack for holding dishes as dishwater drains off of them
\item[mailbox.n.01]	also letter\_box; a private box for delivery of mail
\item[bag.n.04]	also handbag, pocketbook, purse; a container used for carrying money and small personal items or accessories (especially by women)
\item[bicycle.n.01]	also bike, wheel, cycle; a wheeled vehicle that has two wheels and is moved by foot pedals
\item[ladder.n.01]	steps consisting of two parallel members connected by rungs; for climbing up or down
\item[rack.n.05	]also stand; a support for displaying various articles
\item[tray.n.01]	an open receptacle for holding or displaying or serving articles or food
\item[toaster.n.02]	a kitchen appliance (usually electric) for toasting bread
\item[paper\_cutter.n.01]	a cutting implement for cutting sheets of paper to the desired size
\item[plunger.n.03]	also plumber's\_helper; hand tool consisting of a stick with a rubber suction cup at one end; used to clean clogged drains
\item[dryer.n.01]	also drier; an appliance that removes moisture
\item[guitar.n.01]	a stringed instrument usually having six strings; played by strumming or plucking
\item[fire\_extinguisher.n.01]	also extinguisher, asphyxiator; a manually operated device for extinguishing small fires
\item[pitcher.n.02]	also ewer; an open vessel with a handle and a spout for pouring
\item[pipe.n.02]	also pipage, piping; a long tube made of metal or plastic that is used to carry water or oil or gas etc.
\item[plate.n.04]	dish on which food is served or from which food is eaten
\item[vacuum.n.04]	also vacuum\_cleaner; an electrical home appliance that cleans by suction
\item[bowl.n.03]	a dish that is round and open at the top for serving foods
\item[hat.n.01]	also chapeau, lid; headdress that protects the head from bad weather; has shaped crown and usually a brim
\item[rod.n.01]	a long thin implement made of metal or wood
\item[water\_cooler.n.01]	a device for cooling and dispensing drinking water
\item[kettle.n.01]	also boiler; a metal pot for stewing or boiling; usually has a lid
\item[oven.n.01]	kitchen appliance used for baking or roasting
\item[scale.n.07]	also weighing\_machine; a measuring instrument for weighing; shows amount of mass
\item[broom.n.01]	a cleaning implement for sweeping; bundle of straws or twigs attached to a long handle
\item[hand\_blower.n.01]	also blow\_dryer, blow\_drier, hair\_dryer, hair\_drier; a hand-held electric blower that can blow warm air onto the hair; used for styling hair
\item[coatrack.n.01]	also coat\_rack, hatrack; a rack with hooks for temporarily holding coats and hats
\item[teddy.n.01]	also teddy\_bear; plaything consisting of a child's toy bear (usually plush and stuffed with soft materials)
\item[alarm\_clock.n.01]	also alarm; a clock that wakes a sleeper at some preset time
\item[rug.n.01] --merged with mat.n.01--	also carpet, carpeting; floor covering consisting of a piece of thick heavy fabric (usually with nap or pile)
\item[ironing\_board.n.01]	narrow padded board on collapsible supports; used for ironing clothes
\item[fire\_alarm.n.02]	also smoke\_alarm; an alarm that is tripped off by fire or smoke
\item[machine.n.01]	any mechanical or electrical device that transmits or modifies energy to perform or assist in the performance of human tasks
\item[music\_stand.n.01]	also music\_rack; a light stand for holding sheets of printed music
\item[fireplace.n.01]	also hearth, open\_fireplace; an open recess in a wall at the base of a chimney where a fire can be built
\item[furniture.n.01]	also piece\_of\_furniture, article\_of\_furniture; furnishings that make a room or other area ready for occupancy
\item[vase.n.01] an open jar of glass or porcelain used as an ornament or to hold flowers
\item[vent.n.01]	also venthole, vent-hole, blowhole; a hole for the escape of gas or air
\item[candle.n.01]	also taper, wax\_light; stick of wax with a wick in the middle
\item[crate.n.01]	a rugged box (usually made of wood); used for shipping
\item[dustpan.n.02]	a short-handled receptacle into which dust can be swept
\item[earphone.n.01]	also earpiece, headphone, phone; electro-acoustic transducer for converting electric signals into sounds; it is held over or inserted into the ear
\item[jar.n.01]	a vessel (usually cylindrical) with a wide mouth and without handles
\item[projector.n.02]	an optical instrument that projects an enlarged image onto a screen
\item[gat.n.01]	also rod; a gangster's pistol
\item[step.n.04]	also stair; support consisting of a place to rest the foot while ascending or descending a stairway
\item[step\_stool.n.01]	a stool that has one or two steps that fold under the seat
\item[vending\_machine.n.01]	a slot machine for selling goods
\item[coat.n.01]	an outer garment that has sleeves and covers the body from shoulder down; worn outdoors
\item[coat\_hanger.n.01]	also clothes\_hanger, dress\_hanger; a hanger that is shaped like a person's shoulders and used to hang garments on
\item[drinking\_fountain.n.01]	also water\_fountain, bubbler; a public fountain to provide a jet of drinking water
\item[hamper.n.02]	a basket usually with a cover
\item[thermostat.n.01]	also thermoregulator; a regulator for automatically regulating temperature by starting or stopping the supply of heat
\item[banner.n.01]	also streamer; long strip of cloth or paper used for decoration or advertising
\item[iron.n.04]	also smoothing\_iron; home appliance consisting of a flat metal base that is heated and used to smooth cloth
\item[soap.n.01]	a cleansing agent made from the salts of vegetable or animal fats
\item[chopping\_board.n.01]	also cutting\_board; a wooden board where meats or vegetables can be cut
\item[hanging.n.01]	also wall\_hanging; decoration that is hung (as a tapestry) on a wall or over a window
\item[kitchen\_island.n.01]	an unattached counter in a kitchen that permits access from all sides
\item[shirt.n.01]	a garment worn on the upper half of the body
\item[sleeping\_bag.n.01]	large padded bag designed to be slept in outdoors; usually rolls up like a bedroll
\item[tire.n.01]	also tyre; hoop that covers a wheel
\item[toothbrush.n.01]	small brush; has long handle; used to clean teeth
\item[bathrobe.n.01]	a loose-fitting robe of towelling; worn after a bath or swim
\item[faucet.n.01]	also spigot; a regulator for controlling the flow of a liquid from a reservoir
\item[slipper.n.01]	also carpet\_slipper; low footwear that can be slipped on and off easily; usually worn indoors
\item[thermos.n.01]	also thermos\_bottle, thermos\_flask; vacuum flask that preserves temperature of hot or cold drinks
\item[tripod.n.01]	a three-legged rack used for support
\item[dispenser.n.01]	a container so designed that the contents can be used in prescribed amounts
\item[heater.n.01]	also warmer; device that heats water or supplies warmth to a room
\item[pool\_table.n.01]	also billiard\_table, snooker\_table; game equipment consisting of a heavy table on which pool is played
\item[remote\_control.n.01]	also remote; a device that can be used to control a machine or apparatus from a distance
\item[stapler.n.01]	also stapling\_machine; a machine that inserts staples into sheets of paper in order to fasten them together
\item[treadmill.n.01]	an exercise device consisting of an endless belt on which a person can walk or jog without changing place
\item[beanbag.n.01]	a small cloth bag filled with dried beans; thrown in games
\item[dartboard.n.01]	also dart\_board; a circular board of wood or cork used as the target in the game of darts
\item[metronome.n.01]	clicking pendulum indicates the exact tempo of a piece of music
\item[painting.n.01]	also picture; graphic art consisting of an artistic composition made by applying paints to a surface
\item[rope.n.01]	a strong line
\item[sewing\_machine.n.01]	a textile machine used as a home appliance for sewing
\item[shredder.n.01]	a device that shreds documents (usually in order to prevent the wrong people from reading them)
\item[toolbox.n.01]	also tool\_chest, tool\_cabinet, tool\_case; a box or chest or cabinet for holding hand tools
\item[water\_heater.n.01]	also hot-water\_heater, hot-water\_tank; a heater and storage tank to supply heated water
\item[brush.n.02]	an implement that has hairs or bristles firmly set into a handle
\item[control.n.09]	also controller; a mechanism that controls the operation of a machine
\item[dais.n.01]	also podium, pulpit, rostrum, ambo, stump, soapbox; a platform raised above the surrounding level to give prominence to the person on it
\item[dollhouse.n.01]	also doll's\_house; a house so small that it is likened to a child's plaything
\item[envelope.n.01]	a flat (usually rectangular) container for a letter, thin package, etc.
\item[food.n.01]	also nutrient; any substance that can be metabolized by an animal to give energy and build tissue
\item[frying\_pan.n.01]	also frypan, skillet; a pan used for frying foods
\item[helmet.n.02]	a protective headgear made of hard material to resist blows
\item[tennis\_racket.n.01]	also tennis\_racquet; a racket used to play tennis
\item[umbrella.n.01]	a lightweight handheld collapsible canopy
\end{description}

\end{document}